\pdfoutput=1
\documentclass[fleqn,10pt]{wlpeerj}
\usepackage[T1]{fontenc}

\usepackage{amsmath,amssymb}
\usepackage{booktabs}
\usepackage{array}
\usepackage{graphicx}
\usepackage[authoryear,round]{natbib}
\usepackage{hyperref}
\usepackage{orcidlink}
\definecolor{color2}{RGB}{255,255,255}
\usepackage{xurl}
\hypersetup{colorlinks=true,linkcolor=black,citecolor=black,urlcolor=blue!55!black}
\setcitestyle{authoryear,round,aysep={},yysep={;}}
\providecommand{\pkg}[1]{\textbf{#1}}

\title{Machine Learning for Coding Retail Product Names to Consumer-Price Categories: A Rule-plus-Bag-of-Words Pipeline with Reliability-Weighted Human-in-the-Loop Labeling}

\author[1]{Vladimir Beskorovainyi\,\orcidlink{0009-0004-7005-6242}}
\affil[1]{Besk Tech, Moscow, Russia}

\corrauthor[1]{Vladimir Beskorovainyi}{admin@besk.tech}

\begin{abstract}
Price statistics increasingly draw on scanner, web-scraped and receipt data,
whose product descriptions are short, noisy and carry no standard product
code, so each item must be coded to a consumption classification such as
COICOP. That lightweight text classifiers are adequate for this task is
already reported by national statistical offices, but the published evidence
is thin on dispersion, on paired comparison, on train--test overlap and on
measured computational cost. This paper supplies that missing evaluation. On
an openly released synthetic benchmark of six COICOP-like categories, seven
models trained from scratch and two pretrained encoders in three
configurations are run under one matched protocol and compared on both
accuracy and measured cost. A character $n$-gram logistic regression is the
most accurate model in every category (mean $F_1 = 0.997$, and 0.996 on test
items unseen in training) and wins all 30 paired comparisons against each
other model trained from scratch. The small CNN and LSTM examined here do not
have a higher mean $F_1$ than any non-deep model, but once trained to the same
stopping criterion as everything else they are not reliably worse than the
word-level models either; what separates them is cost, at 212 and 258 times
the training time of the cheapest model for no gain in accuracy. A fine-tuned
DistilBERT comes closest to the character model ($F_1 = 0.993$) without
passing it, taking 3 of 30 paired comparisons to the character model's 18
with 9 ties, at about 460 times its inference cost on CPU with PyTorch limited
to one thread; frozen
MiniLM sentence embeddings do no better than bag-of-words. The most accurate
model is not, however, the cheapest: character features cut inference
throughput to 41{,}500 items per second against 191{,}300 for unigram
bag-of-words, a measurement the informal version of the claim does not
supply. A rule-based prefix-tree stage admits 75--86\% of positive items,
so in a cascade it bounds the recall of the whole pipeline. A Monte Carlo study of the labeling
protocol, in which annotators are simulated and no human annotation was
collected, shows that an additive reliability weight barely improves on
majority vote while Dawid--Skene aggregation recovers labels markedly better,
and that this difference carries through to the character $n$-gram classifier trained on the
aggregated labels. The benchmark is synthetic and English-language because the
production data behind the architecture are subject to non-disclosure; its
generator applies character-level corruption, plants a grade token in most
positive items, and shares phrase sets with the rule stage, and these
conditions are stated, and where possible tested, wherever a result depends on
them.
\end{abstract}

\begin{document}

\flushbottom
\maketitle
\thispagestyle{empty}

\section{Introduction}
\label{sec:intro}

The Consumer Price Index is among the most consequential economic statistics
produced by a state: it drives monetary policy, the indexation of wages and
pensions, and the deflation of national accounts. Traditionally its price
quotes are collected by field staff who visit outlets and record prices for a
fixed basket of representative products, a process that is costly, limited in
spatial and temporal coverage, and burdensome for respondents.

Digital transaction records offer an alternative. Retail scanner data and
web-scraped prices have been studied for over a decade as inputs to price
statistics \citep{deHaanGrient2011, Chessa2016, CavalloRigobon2016,
Cavallo2017, Bialek2021}. A complementary and less explored source is
transaction and receipt microdata, where networked cash registers record
sales at high frequency. Such records give a broad view of retail prices and
quantities, but their item descriptions are short, abbreviated and
inconsistently formatted; they frequently contain concatenated tokens, carry
brand and quantity information in free text, and lack a standardized product
code. Before such data can feed an index, each line item must be mapped to a
consumption classification such as COICOP \citep{COICOP} and its unit and
pack size normalized.

How much model complexity that mapping requires is, in one sense, a settled
question. \citet{HarmsSpinder2019} compared logistic regression, naive Bayes,
random forests, support vector machines and gradient boosting against
recurrent, convolutional and attention architectures on real product
descriptions for the Dutch CPI, and found logistic regression both the most
accurate and by far the fastest, attributing the result to descriptions too
short and non-cohesive for deep models to exploit. \citet{Oancea2023} reached
a compatible conclusion on real ECOICOP data across thirteen classifiers and
nine representations. Several statistical offices report operational systems
built on exactly this premise \citep{ONSscanner, ONSnewdata2021, Jentoft2022,
Montbroussous2024}.

What that literature does not supply is the evidence a methodologist needs in
order to rely on the premise. Dispersion across random splits is rarely
reported; comparisons are made between point estimates rather than by paired
test; train--test overlap in the evaluation corpus is not audited; the
rule-based stages that precede the classifier in real pipelines are not
separately evaluated; and the computational cost on which the
argument turns is almost never measured. This paper supplies that evaluation. Its
contribution is not the finding that lightweight models suffice, which is
already established, but a reproducible protocol under which the finding can
be checked, extended and contradicted, together with the cost measurement
that the informal claim lacks.

The architecture under evaluation is a hybrid: a deterministic prefix-tree
screening stage configured by a methodologist, followed by a per-category
binary confirmation model, with training labels produced by a
human-in-the-loop protocol that aggregates annotator votes under a
reliability weight. That architecture was developed for, and has been
operated in, a continuously running production environment in a
non-English-language deployment, processing tens of millions of line items per
day, which is why per-item inference cost is treated here as a first-class
criterion. Because the data of that deployment are subject to non-disclosure, the production experience
is reported here as context for the design and never as evidence for a
quantitative claim. All numbers in this paper come from a synthetic benchmark
released in full, and the annotators in the labeling study are simulated: no
human annotation was collected for this work.

Three questions organize what follows. First, how do model families compare
on short noisy product names when the protocol is matched, dispersion is
reported, comparisons are paired, and the corpus is audited for overlap?
Second, what does each model actually cost, in training time, inference
throughput and serialized size, on identical hardware? Third, does the choice
of label-aggregation rule matter not only for recovering labels but for the
accuracy of the classifiers subsequently trained on them? The answers are, in
order: a character $n$-gram logistic regression wins every paired comparison
against the other models trained from scratch, while neither of the small deep
models examined here has a higher mean $F_1$ than any non-deep model, though
they are not reliably beaten by the word-level models once their training
budget is matched; the most accurate
model is not the cheapest, and the accuracy--cost frontier is occupied by
three models, with the deep ones off it entirely; and the aggregation rule
does carry through to the character model, with a transparent additive
weight performing close to majority vote and a latent-ability estimator
clearly better than both. Pretrained encoders, added under the same protocol,
leave that ordering in place: a fine-tuned DistilBERT beats every model except
the character one, which it approaches without passing (the character model
takes 18 of the 30 paired cells, DistilBERT 3, with 9 ties), at several
hundred times its inference cost.

\section{Related work}
\label{sec:related}

\subsection{Alternative data for price statistics}

Scanner data, that is transaction records from retailer point-of-sale
systems, have been incorporated into several national indices, motivating new
index methodology to handle high product churn and to exploit quantity
weights \citep{deHaanGrient2011, Chessa2016}. \citet{Bialek2021} set out the
full pipeline from raw scanner records to price indices, with classification
to a low COICOP level as one of its stages. The Billion Prices Project and
related work showed that large-scale online price data can track and even
anticipate official inflation \citep{CavalloRigobon2016, Cavallo2017}.
Eurostat's practical guide for processing supermarket scanner data states the
coding problem directly, naming supervised learning on item descriptions as
the route to take where a retailer's own classification is unavailable
\citep{EurostatScanner2017}. Transaction and receipt data of the kind
considered here share the unstructured-description problem of scanner data
but can offer broader outlet coverage, at the cost of weaker provenance
metadata.

\subsection{Automated coding in official statistics}

Coding free text to a statistical classification is long-standing practice in
statistical offices, and the recent literature is specific enough to
calibrate against. \citet{ONSscanner} compared five classical models across
85 clothing consumption segments on real web-scraped data, with the best
reaching macro-averaged precision near 0.79 against a human-labeller
consistency ceiling of 88.8\%, well below the figures a controlled corpus
produces. \citet{Martindale2020}, also at the UK office, attacked the scarce-label
problem directly, propagating labels from a small seed set with word
embeddings and fuzzy matching. \citet{ONSnewdata2021} set out the
operational design space, distinguishing fully automatic coding, automatic
coding above a confidence threshold, and machine-assisted recommendation to
human labellers. \citet{Jentoft2022} compared four classical models on
Norwegian supermarket transaction data and proposed a confidence-threshold
loop with human review, closely related to the protocol evaluated in
Section~\ref{sec:labeling}. \citet{Montbroussous2024} report a wider gap still: a model reaching 97.4\% on pre-labelled test data coded only
41.4\% of expenditure correctly at six-digit COICOP on genuinely unlabelled
hard-discount data, which is the clearest published statement of the gap
between held-out accuracy and production accuracy on this task.
\citet{Schierholz2021} compared seven algorithms for occupation coding across
five datasets and concluded that algorithm choice matters less than the
characteristics of the data and of the original coding, a finding whose
analogue for product coding this paper's results do not contradict. At the
current frontier, \citet{Toth2025} evaluates self-hosted large language
models with retrieval for COICOP coding against BERT classifiers and human
coders, and \citet{Berki2025} report 94.6\% accuracy with BERT on consumer
electronics; \citet{AIML4OS2024} describes the European Statistical System's
current work package on coding free text to NACE, COICOP and ISCO.
Hierarchical structure in the target classification is itself a modelling
choice: \citet{Beresewicz2024} show that hierarchy-aware decoding buys one to
two percentage points on occupation coding.

Two studies bear directly on this paper's model comparison and are therefore
treated as its baseline rather than as background. \citet{HarmsSpinder2019}
benchmarked classical models against recurrent, convolutional and attention
architectures on real CPI product descriptions and found logistic regression
best, at 92.1\% accuracy in 43 seconds against more than nine hours for a
feed-forward network. Their evaluation covers a single clothing vendor and
targets that vendor's own website categories rather than COICOP, because
manually labelled data were not yet available to them; dispersion, paired
tests and overlap auditing are not reported, and the data and code are not
released. \citet{Oancea2023} crossed thirteen classifiers with nine
representations on 2{,}853 products in fifteen ECOICOP classes, found
logistic regression, support vector machines and random forests best in
combination with fastText skip-gram embeddings, and reported total
computation time, though for the study as a whole rather than per model; the
corpus is small and heavily imbalanced, and robustness across splits is not
analysed. The present work differs in what it measures rather than in what it
concludes.

\subsection{Quality, drift and governance}

Because a coding model is an intermediate rather than a final statistical
output, its quality framework differs from that of a published estimate.
\citet{Yung2022} present the UNECE quality framework for statistical
algorithms, designed for exactly such intermediate outputs, and
\citet{Molladavoudi2023} operationalise two of its dimensions, including
continuous monitoring of auto-coding. \citet{Meertens2022} treat concept
drift as the central obstacle to machine learning in official statistics and
compare estimators that correct the resulting bias. These concerns motivate
the continual-learning loop of Section~\ref{sec:labeling} but are not
themselves evaluated here.

\subsection{Learning from noisy labels}

Where ground truth comes from several imperfect annotators, aggregating their
judgments and estimating their reliability is a classical problem.
\citet{DawidSkene1979} introduced an EM procedure that jointly estimates true
labels and per-annotator error rates; later work studied repeated labeling
and its effect on model quality \citep{Sheng2008}, the value of non-expert
annotation \citep{Snow2008}, and models that estimate annotator ability
jointly with item labels or with the classifier itself
\citep{Whitehill2009, Raykar2010}. The reliability-weighted rule evaluated in
Section~\ref{sec:consensus-sim} is a lightweight online variant, adopted in
production because an audited scalar score per assessor was an operational
requirement, and it is assessed here as one design choice among three rather
than as a contribution.

\subsection{Text representation}

Mapping short product strings to a classification is a short-text
classification problem, for which distributed word representations and
subword models \citep{Mikolov2013, Bojanowski2017}, efficient linear
classifiers over such features \citep{Joulin2017}, and pretrained Transformer
encoders \citep{Devlin2019} are the standard tools. The models compared in
Section~\ref{sec:eval} are deliberately drawn from the cheap end of that
range, for reasons Section~\ref{sec:discussion} returns to.

\section{Problem setting and architecture}
\label{sec:setting}

\subsection{Input, targets and stages}

The input is a feed of two record types typical of any transaction or receipt
source: registration records, one per point-of-sale device, carrying location
and economic-activity codes, and line-item records, one per receipt line,
carrying a free-text item name, quantity, unit price and line sum. The two are
joined on the device identifier, and records failing the join or integrity
checks are quarantined rather than dropped, so that coverage remains
measurable. Each retained line item must be assigned to a node of the
consumption classification, cross-walked to the index item-representative
list, and its quantity and pack size normalized, for instance to a price per
kilogram, because the unit of analysis is a price quote for a representative
product in an outlet--region--period cell.

\begin{figure}[hbt]
\begin{center}
\includegraphics[width=\linewidth]{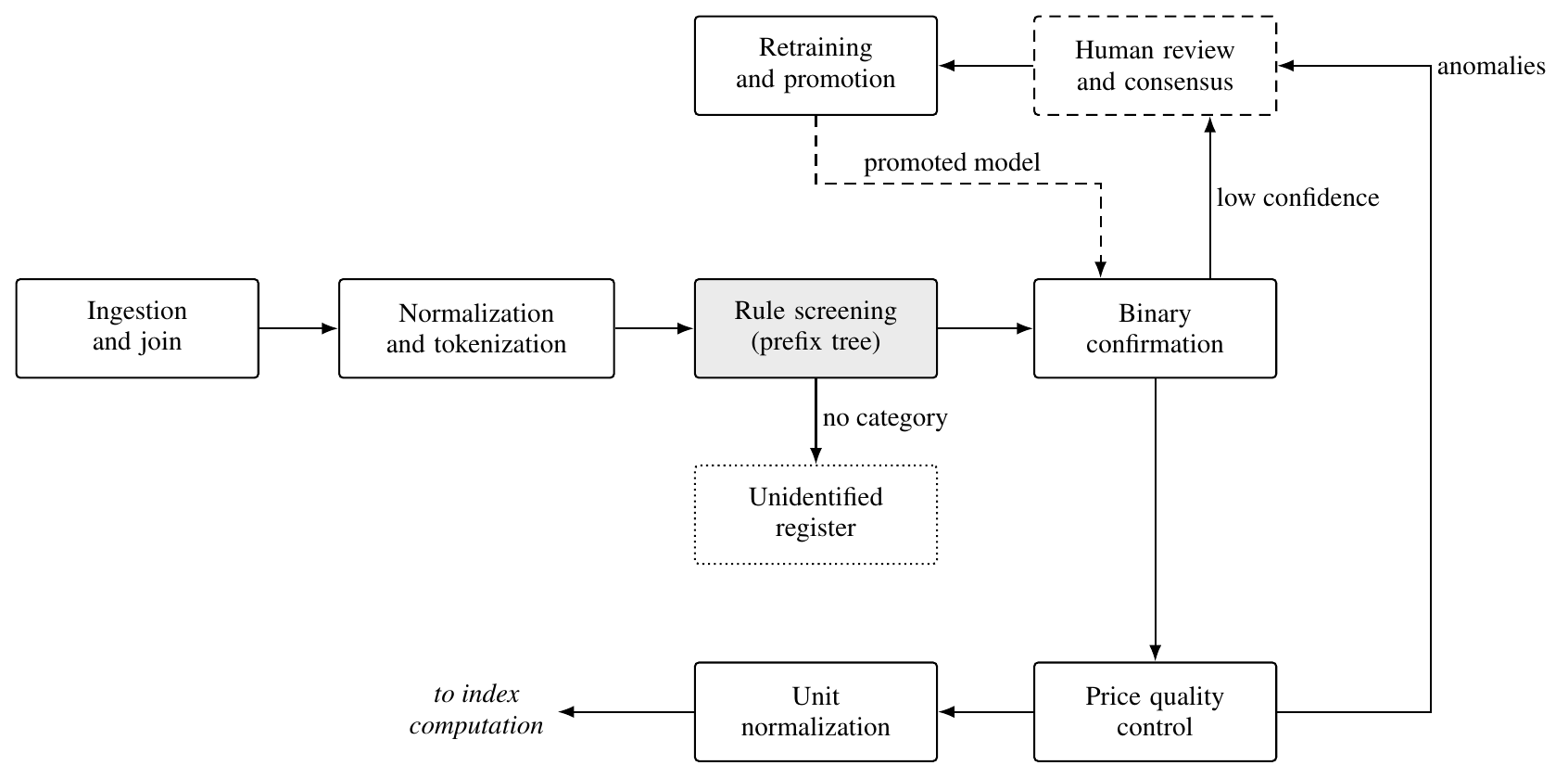}
\end{center}
\caption{\label{fig:arch}Stages an item passes through. Solid arrows carry
items, dashed arrows carry feedback. The shaded stage is deterministic and
configured by a methodologist, the dashed stage is performed by people, and
the dotted stage is where items the rules admit to no category leave the
flow; items admitted by more than one category go to expert review. Storage and the correction journal underlie every stage and are not
drawn. This is a schematic of the architecture, not a result.}
\end{figure}

The method is a sequence of stages: ingestion; product identification,
comprising normalization, tokenization, rule screening and per-category
confirmation; labeling and learning; quality control; and aggregation of
confirmed quotes for downstream index computation, with raw data and an
auditable correction journal underlying all of them. Figure~\ref{fig:arch}
draws the stages an item passes through. For labeling it is useful to
separate three roles: annotators who label items, expert assessors whose
judgments carry more weight, and methodologists who own the category
definitions.

\section{Materials and methods}
\label{sec:methods}

Identification proceeds in three stages: normalization and tokenization, rule
screening, and per-category binary confirmation. Training labels for the
third stage are produced by the protocol of Section~\ref{sec:labeling}.

\subsection{Normalization and rule screening}
\label{sec:trie}

Normalization removes semantically empty characters and tokens while
preserving the attributes that matter for price comparability, such as fat
content, alcohol strength, brand, grade, pack size and unit. Tokenization
additionally attempts to split concatenated tokens; names that cannot be split
are routed to an unidentified register for review rather than silently
discarded.

Each target category is configured by a methodologist with a set of key
phrases that act as positive triggers and a set of stop phrases that act as
exclusions, matched against tokenized item names with a prefix tree. For an item with token sequence $x$ and a
category $c$ with key-phrase set $K_c$ and stop-phrase set $S_c$, the stage
admits $x$ to $c$ if and only if
\begin{equation}
\Big(\exists\, p \in K_c : p \sqsubseteq x\Big)
\;\wedge\;
\Big(\nexists\, q \in S_c : q \sqsubseteq x\Big),
\label{eq:trie}
\end{equation}
where $p \sqsubseteq x$ denotes that phrase $p$ occurs as a contiguous token
subsequence of $x$. Items matched uniquely are passed to confirmation. Items admitted by no
category are written to the unidentified register, which is available for
review but does not feed further processing. Items admitted by more than one
category are not resolved by the rules: they are routed to expert review, in
the same queue as the abstentions and cross-category indications of
Section~\ref{sec:labeling}. In the benchmark of Section~\ref{sec:eval} no item
is admitted by more than one category (\path{results/coverage.csv}), so the
case does not arise there, and
Section~\ref{sec:discussion} notes multi-category assignment as outside its
scope. The stage is therefore a filter, and its recall on positive items
bounds the recall of the pipeline, as Section~\ref{sec:trieresults}
quantifies.

\subsection{Per-category binary confirmation}
\label{sec:confirm}

The residual decision is framed as binary confirmation: given a tokenized
item that screening has tentatively assigned to a category $c$, a model $f_c$
decides whether the assignment is valid or should be rejected. This framing
mirrors the judgment human assessors make, so machine and human labels are
interchangeable and can be aggregated by the same rule. A separate model is
maintained per category, so that a category can be added or retrained
without disturbing the rest, which matters for an index whose basket changes
over time.

Candidate forms for $f_c$ span a deliberately narrow and cheap range.
Item names are encoded as binary feature vectors over a vocabulary fitted on
training data; the classifier is then a logistic regression over unigram,
word $1$--$2$-gram, word $1$--$3$-gram or character $3$--$5$-gram features, or
a multilayer perceptron with one hidden layer over unigram features. Two
order-sensitive alternatives are included for comparison: a one-dimensional
convolutional network and an LSTM, each over learned embeddings of
whitespace-delimited tokens. Table~\ref{tab:hyper} lists every setting. Because
the CNN and LSTM vocabularies are built from whole tokens, a corrupted token
unseen in training maps to a single unknown-token symbol, whereas the
character model still sees most of its $n$-grams; the comparison between the
deep and the shallow models is therefore also a comparison between
representations, and character-aware neural models \citep{Zhang2015charcnn}
were not tested. Discarding word order is plausible here because item names
are short and bag-of-words cues such as brand, grade and unit are highly
discriminative; Section~\ref{sec:eval} tests that expectation rather than
assuming it.

\begin{table}[hbt]
\caption{\label{tab:hyper}Model settings, as implemented in
\texttt{src/confirm\_models.py} and, for the pretrained encoders, in
Supplemental File~1. Settings not listed are library defaults; none was tuned
on test results.}
\begin{center}\small
\begin{tabular}{>{\raggedright\arraybackslash}p{0.2\linewidth}>{\raggedright\arraybackslash}p{0.72\linewidth}}
\toprule
Model & Settings \\
\midrule
Logistic regression (four feature sets) & L2 penalty, $C = 1$, L-BFGS solver, at most 1{,}000 iterations; binary \texttt{CountVectorizer} over word $1$-, $1$--$2$- or $1$--$3$-grams, or \texttt{char\_wb} $3$--$5$-grams; vocabulary fitted on the training split \\
MLP & Unigram binary features; one hidden layer of 256 units; Adam with \pkg{scikit-learn} defaults; at most 80 epochs, stopping when the training loss fails to improve by more than $10^{-4}$ for 10 consecutive epochs \\
CNN & Token embedding of dimension 32; 64 convolutional filters of width 3; global max-pooling; linear output \\
LSTM & Token embedding of dimension 32; one LSTM layer with 64 hidden units; linear output on the final hidden state \\
CNN and LSTM & Adam, learning rate $2\times10^{-3}$, batches of 128, binary cross-entropy loss; vocabulary of up to 4{,}000 whitespace tokens built on the training split, plus padding and unknown symbols; inputs truncated or padded to 12 tokens; stop when the training loss fails to improve by more than $10^{-4}$ for 10 consecutive epochs, at most 200 epochs \\
DistilBERT & \texttt{distilbert-base-uncased}, fine-tuned end to end with a two-class head; AdamW, learning rate $2\times10^{-5}$, batches of 32, at most 32 subword tokens; stopping as for the CNN and LSTM, or a fixed 3 epochs \\
MiniLM & \texttt{all-MiniLM-L6-v2}, frozen; mean-pooled, unit-normalized sentence embeddings; logistic regression as above \\
\bottomrule
\end{tabular}
\end{center}
\end{table}

\subsection{Human-in-the-loop labeling}
\label{sec:labeling}

After screening, labeling tasks grouped by category are dispatched to
annotators and expert assessors, who see each item's name and price and give
a judgment guided by the category's annotation instruction. An annotator may
also abstain, or indicate that the item belongs to a different category; such
items are routed to expert review and do not enter the vote.

\subsubsection{The consensus rule}

Each annotator $a$ carries a bounded reliability weight $r_a \ge 0$, updated
automatically over time. Encode annotator $a$'s vote on an item as
$v_a \in \{+1,-1\}$, for valid and reject respectively. The aggregate opinion
is the reliability-weighted sum of votes,
\begin{equation}
O \;=\; \sum_{a \in A} r_a \, v_a,
\qquad
\hat{y} \;=\;
\begin{cases}
\text{valid}, & O > 0,\\
\text{reject}, & O < 0,\\
\text{defer to human adjudication}, & O = 0,
\end{cases}
\label{eq:consensus}
\end{equation}
where $A$ is the set of assessors who voted on the item. In deployment ties
are deferred rather than decided automatically. The simulation of
Section~\ref{sec:consensus-sim} has no adjudicator to defer to, so it
implements the coin-flip variant: a tie is broken at random and the weights
are then updated on that outcome as on any other.

\subsubsection{The weight update}
\label{sec:weights}

The update rule is specified here in full, since
Section~\ref{sec:consensus-sim} evaluates it. Every annotator enters with
weight $r_a = r_0$. After the consensus $\hat{y}$ for an item is formed, each
assessor who voted on that item has their weight adjusted additively
according to whether their vote agreed with the consensus,
\begin{equation}
r_a \;\leftarrow\;
\begin{cases}
\min\!\left(r_{\max},\, r_a + \delta\right), & v_a = \operatorname{sign}(O),\\[2pt]
\max\!\left(0,\, r_a - \delta\right), & v_a \neq \operatorname{sign}(O),
\end{cases}
\label{eq:update}
\end{equation}
where $\operatorname{sign}(O) \in \{+1,-1\}$ is the sign of the aggregate
opinion, which is the consensus $\hat{y}$ expressed on the same scale as the
votes, and with $r_0 = 1$, $\delta = 0.1$ and $r_{\max} = 2$, the values used
in production and in the simulation below. Four points complete
the specification. Updates are applied item by item in arrival order, so the
rule is online and a weight depends on the order in which items were
adjudicated. An abstention is not a vote: the annotator's weight is left
unchanged, since there is nothing to agree or disagree with. In deployment a
deferred item ($O = 0$) updates no weights, because no consensus was formed;
in the simulation the randomly broken tie is treated as the consensus. Once a
category's model passes its quality gate its predictions enter
Equation~\ref{eq:consensus} as an additional assessor with the same initial
weight $r_0$ and the same update rule, so that machine and human judgments
are aggregated uniformly; the model is not a voter in the simulation of
Section~\ref{sec:consensus-sim}, which isolates the aggregation rule, and
Section~\ref{sec:discussion} returns to the feedback risk this creates.

Note that Equation~\ref{eq:update} saturates. The weight of any annotator who
agrees with the consensus more often than not is a random walk with positive
drift against a reflecting barrier at $r_{\max}$, so it concentrates there,
which is the mechanism behind the result in Section~\ref{sec:consensus-sim}.

Three further stages complete the pipeline and are not evaluated in this
paper: continual learning, in which a category's model is retrained once its
newly labeled items cross a configured threshold and is promoted only if it
passes re-validation on held-out items; quality control, comprising integrity
checks, coverage metrics and price-bound control; and the aggregation of
confirmed quotes into indices \citep{CPIManual}.

\subsection{A reproducible synthetic benchmark}
\label{sec:benchmark}

The production data behind the architecture are subject to non-disclosure and
cannot be released, which would leave any evaluation on them unverifiable.
The evaluation is therefore conducted on a synthetic, English-language
benchmark of six COICOP-like categories, with UK-style brand names: granulated
sugar, milk, bread, beer, laundry detergent and fresh apples. For each category a generator produces 700
positive and 950 hard-negative receipt-style names from fixed seeds, with
abbreviations, missing spaces, character transpositions and distractors that
share tokens with the target without being it, such as icing sugar,
sugar-free yogurt, coconut milk and milk chocolate. Appendix~\ref{app:gen}
specifies the generator in full. All data, models and scripts are released as
a reproducibility package at
\url{https://github.com/beskvladimir-create/cpi-coding-repro}. Every accuracy
figure except those in the pretrained-encoder sections is reproduced exactly
by release 2.1.1 of that package, archived on Zenodo under DOI
\texttt{10.5281/zenodo.23445113}; timings are reproducible as ratios rather
than as absolute rates; and the pretrained-encoder results come from
Supplemental File~1.

What a synthetic corpus can and cannot establish is stated in
Section~\ref{sec:discussion} and carried into every claim below. Three
properties matter throughout: the generator applies character-level
corruption, which creates conditions favourable to character $n$-gram
features; it gives most positive items a grade token that no negative item
carries, a shortcut whose effect Appendix~\ref{app:grade} measures; and the
rule stage of Section~\ref{sec:trie} consumes the same phrase sets the
generator draws on, so its measured coverage is an optimistic bound.

\subsection{Protocol}
\label{sec:protocol}

All seven models are trained under one matched protocol: a stratified
$80/20$ split per seed, five seeds, and the vocabulary fitted on the training
split only, since building it over the full corpus before splitting leaks
test information. Reported quantities are the $F_1$ of the positive class at
the default decision threshold of $0.5$; Appendix~\ref{app:robust} repeats the
comparison under a threshold-free metric.

Matching the protocol includes matching when training stops, which is easy to
get wrong. Every model here runs to its own convergence criterion rather than
for a fixed number of passes: the logistic models stop on the solver's
tolerance, reaching it after a median of 14 to 16.5 of their 1{,}000 permitted
iterations, and the multilayer perceptron and the two deep models stop when
the training loss fails to improve by more than $10^{-4}$ over ten consecutive
epochs. The version of this benchmark reported in an earlier version (v3) of
the preprint \citep{Beskorovainyi2026arxiv} instead trained the convolutional and recurrent models for a fixed eight
epochs while every other model converged, which cost the CNN 1.6 $F_1$ points
and the LSTM 1.5 and was enough to change the conclusion drawn from the
comparison. Appendix~\ref{app:robust} reports that sensitivity. Under the
present rule the deep models stop after a median of 47.5 epochs, 54 for the CNN
(range 46 to 62) and 36 for the LSTM (28 to 60), against a cap of 200, and the
multilayer perceptron after a median of 73 of its cap of 80. None of the 210
model fits reached its iteration limit and all converged
(\path{results/robustness_convergence.csv}); the released code no longer
suppresses convergence warnings, so a future run that did not converge would
say so.

Differences between models are summarized by effect size and by consistency
across the 30 category--seed cells, and the text reports win, loss and tie
counts over those cells throughout. Paired two-sided Wilcoxon signed-rank
tests \citep{Wilcoxon1945}, Holm-corrected across model pairs
\citep{Holm1979}, were computed and are listed in
Appendix~\ref{app:pairs}, but no conclusion here rests on a $p$-value,
because the five seeds within a category are
repeated splits of one synthetic corpus rather than independent
replications. Tests over overlapping resamples of one dataset are known to
understate the variance of the difference and to inflate type~I error
\citep{Dietterich1998, NadeauBengio2003}, and the multiple-dataset procedures
of \citet{Demsar2006} and \citet{Benavoli2017} presuppose independent
datasets, which six categories from one generator are not. The text therefore
uses three fixed bands on the share of non-tied cells that the model with the
higher mean wins against the other: a difference is called reliable when
that model wins 90\% or more of the non-tied cells, small but consistent when
it wins at least 70\% but less than 90\%, and the two models are called
indistinguishable when it wins less than 70\%, or when fewer than 10 of the
30 cells are non-tied, too few to establish a direction. Ties are left out of
the share because near the ceiling many cells are tied, often at $F_1 = 1$,
and say nothing about which model is ahead. The bands describe how
consistent the direction of a difference is, not how large it is, so the mean
difference is always given beside them. The resolution
of a cell is coarse: each test split holds 140 positives, so a single error
moves $F_1$ by about 0.36 points near the ceiling, and differences of a few
tenths of a point correspond to about one item per cell.

Computation ran on one CPU thread of an Intel Xeon at 2.80\,GHz, under
Python 3.13.16 \citep{Python} with \pkg{scikit-learn} 1.9.1 \citep{sklearn},
\pkg{NumPy} 2.2.6 \citep{numpy}, \pkg{SciPy} 1.17.1 \citep{scipy},
\pkg{pandas} 2.2.3 \citep{pandas} and \pkg{PyTorch} 2.5.1 \citep{pytorch}.
The single thread is enforced rather than assumed: the BLAS and OpenMP thread
pools are limited to one through \pkg{threadpoolctl} and the corresponding
environment variables, PyTorch is set to one intra-op and one inter-op
thread, and the thread-pool state during measurement is recorded in the
released results.

\subsection{Pretrained encoder baselines}
\label{sec:pretrained-methods}

The seven models above are trained from scratch. To test whether pretrained
weights change the comparison, two pretrained encoders were run in three
configurations under the same protocol: the same six corpora read from
release 2.0.1, whose data are identical to those of release 2.1.1, the same stratified $80/20$ splits with seeds 0 to 4, the same
metrics, and $F_1$ on test items unseen in training computed as in
Appendix~\ref{app:overlap}.

The first is DistilBERT \citep{Sanh2019}, \texttt{distilbert-base-uncased} at
revision \texttt{12040ac}, fine-tuned end to end with AdamW
\citep{Loshchilov2019} under the stopping rule the CNN and LSTM use, with the
settings of Table~\ref{tab:hyper}. The second is the same model fine-tuned for a fixed three
epochs, the usual recipe, reported so that no conclusion rests on the
stopping rule alone. The third uses frozen sentence embeddings from
\texttt{all-MiniLM-L6-v2} \citep{Wang2020MiniLM, Reimers2019} at revision
\texttt{1110a24}, with a logistic regression on top. The
learning rate, batch size and input length are standard settings for this
model and were not tuned: no learning-rate or epoch search was run, no
validation split was used for stopping, and no setting was chosen by looking
at test results. The fixed-three-epoch variant bounds the effect of the
stopping rule but not of the other settings, so the DistilBERT figures are
those of a standard recipe rather than of a tuned model.

Fine-tuning and embedding ran on one NVIDIA Tesla T4 GPU in a Kaggle notebook,
under Python 3.13, \pkg{PyTorch} 2.11 \citep{pytorch} and \pkg{Transformers}
5.16 \citep{Wolf2020}. GPU training is not bit-deterministic, so a rerun
reproduces these $F_1$ values closely but not exactly. Training times on a GPU
are not comparable with the single-thread figures of Table~\ref{tab:cost} and
are reported only for their order of magnitude. Inference cost, which is what
a statistical office pays for every item it codes, was measured on the CPU of
the Kaggle host, following the protocol of Appendix~\ref{app:cost}, with
PyTorch limited to one thread, tokenization included, after a warm-up, as the
fastest of three passes over the test split of seed 0 in each category. The
tokenizer's own thread pool was not limited. Because that host differs from
the one behind Table~\ref{tab:cost}, the unigram and character $n$-gram
logistic models were
re-timed in the same session, with numerical libraries limited to one thread,
and the pretrained models are compared with them through ratios rather than
absolute rates. The script, the Kaggle notebook that ran it, the cost file it
wrote, and the per-run $F_1$ values, per-category precision and recall and
environment record, transcribed from the notebook's run log and output
preview at the precision displayed, are provided as Supplemental File~1.

\section{Results}
\label{sec:eval}

\subsection{Model comparison}
\label{sec:models}

\begin{figure}[hbt]
\begin{center}
\includegraphics[width=0.92\linewidth]{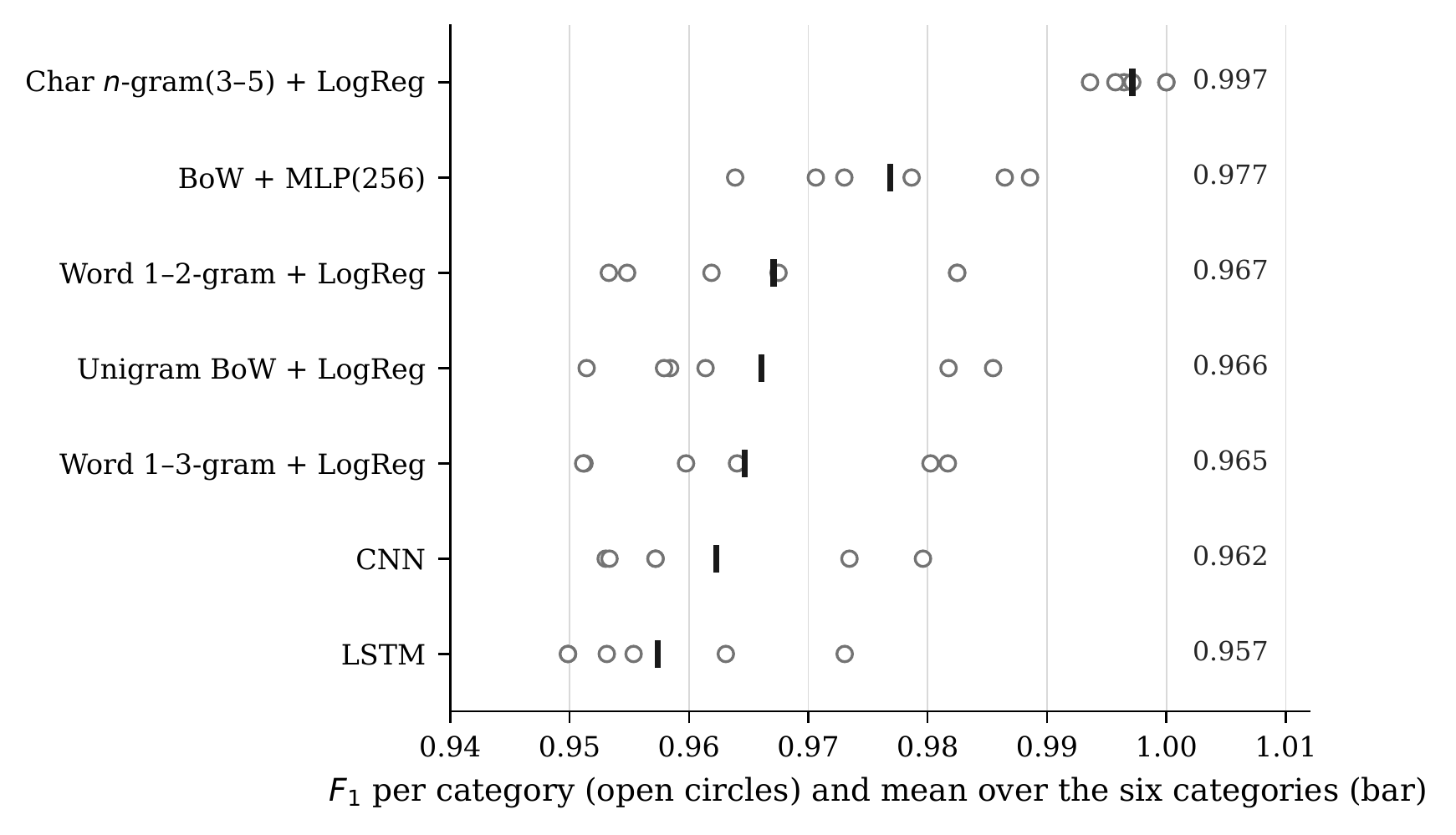}
\end{center}
\caption{\label{fig:ranking}$F_1$ by model under the matched protocol. Open
circles are the six category means and the vertical bar is their average.
The spread of the circles is the spread across categories; it is not a
confidence interval.}
\end{figure}

Table~\ref{tab:models} and Figure~\ref{fig:ranking} report $F_1$ across the
six categories. Three results stand out.

A character $n$-gram logistic regression is the most accurate of the seven
models, at mean $F_1 = 0.997$, and it wins all 30 paired comparisons against
each of the other six. Its margin over the next model is 2.0 $F_1$ points and
over the weakest 4.0 points, and it is the only model whose worst category
exceeds the best category of every other model trained from scratch.

Word order contributes little. Unigram features are indistinguishable from
word $1$--$2$-grams, taking 10 of the 30 cells against 12 with 8 ties, and
from word $1$--$3$-grams, taking 17 against 9 with 4 ties; the three
word-level feature sets lie within 0.3 $F_1$ points of one another. The
$1$--$2$-gram model never loses a cell to the $1$--$3$-gram model, taking 17
with 13 ties, which the bands call reliable; the difference is consistent in
direction but small in size, 0.2 points on average, less than one item per
cell. The
multilayer perceptron is ahead of each word-level logistic model by a small
but consistent margin, taking 23 to 25 of the 30 cells, a lead that
Appendix~\ref{app:noise} finds to depend on the generator's character-level
noise.

Neither deep model has a higher mean $F_1$ than any non-deep model, but
neither clearly loses to the word-level models. The CNN reaches 0.962 and the
LSTM 0.957, placing both below every other model on the mean. Per cell the
picture is weaker than that ordering suggests: the word-level logistic models
take 17 or 18 of the 30 cells against the CNN, which takes 10 to 12, and 18 to
20 against the LSTM, which takes 9 to 11; all six pairs fall inside the band
called indistinguishable, the word-level model winning 59\% to 69\% of the
non-tied cells, with the LSTM against the word $1$--$2$-gram model
(9 cells to 20, one tie) at its edge. Only against the character $n$-gram model
and the multilayer perceptron are they reliably beaten: the character model
takes all 30 cells against each of them, and the MLP takes 27 of 30 against
the CNN and 28 of 30 against the LSTM. Between the two deep models the CNN is
ahead by a small but consistent margin, taking 21 cells to the LSTM's 8 with
one tie, for a mean difference of 0.5 points.

The defensible statement is therefore that on this benchmark the deep models
buy nothing, not that they are worse. They are small, lightly tuned
architectures trained on a few hundred examples per category, so the
conclusion is about architectures of that kind in this regime and does not
extend to deep learning generally. Pretrained encoders, which behave
differently, are examined separately in Section~\ref{sec:pretrained}.

\begin{table}[hbt]
\caption{\label{tab:models}$F_1$ by model under the matched protocol of
Section~\ref{sec:protocol}. The mean is over the six categories; the seed
standard deviation is the sample standard deviation ($n-1$) across the five
seeds within a category, averaged over categories, from
\texttt{results/significance\_dispersion.csv}; the range is the spread of the six category
means and is not an uncertainty interval.}
\begin{center}
\begin{tabular}{lccc}
\toprule
Model & Mean $F_1$ & Seed s.d. & Range over categories \\
\midrule
Char $n$-gram ($3$--$5$) + logistic reg. & $\mathbf{0.997}$ & $0.002$ & $0.994$--$1.000$ \\
Unigram bag-of-words + MLP ($256$) & $0.977$ & $0.008$ & $0.964$--$0.989$ \\
Word $1$--$2$-gram + logistic reg. & $0.967$ & $0.014$ & $0.953$--$0.982$ \\
Unigram bag-of-words + logistic reg. & $0.966$ & $0.014$ & $0.951$--$0.985$ \\
Word $1$--$3$-gram + logistic reg. & $0.965$ & $0.014$ & $0.951$--$0.982$ \\
CNN (embedding + convolution) & $0.962$ & $0.011$ & $0.953$--$0.980$ \\
LSTM (embedding + recurrence) & $0.957$ & $0.012$ & $0.950$--$0.973$ \\
\bottomrule
\end{tabular}
\end{center}
\end{table}

\subsection{Computational cost}
\label{sec:cost}

\begin{figure}[hbt]
\begin{center}
\includegraphics[width=0.95\linewidth]{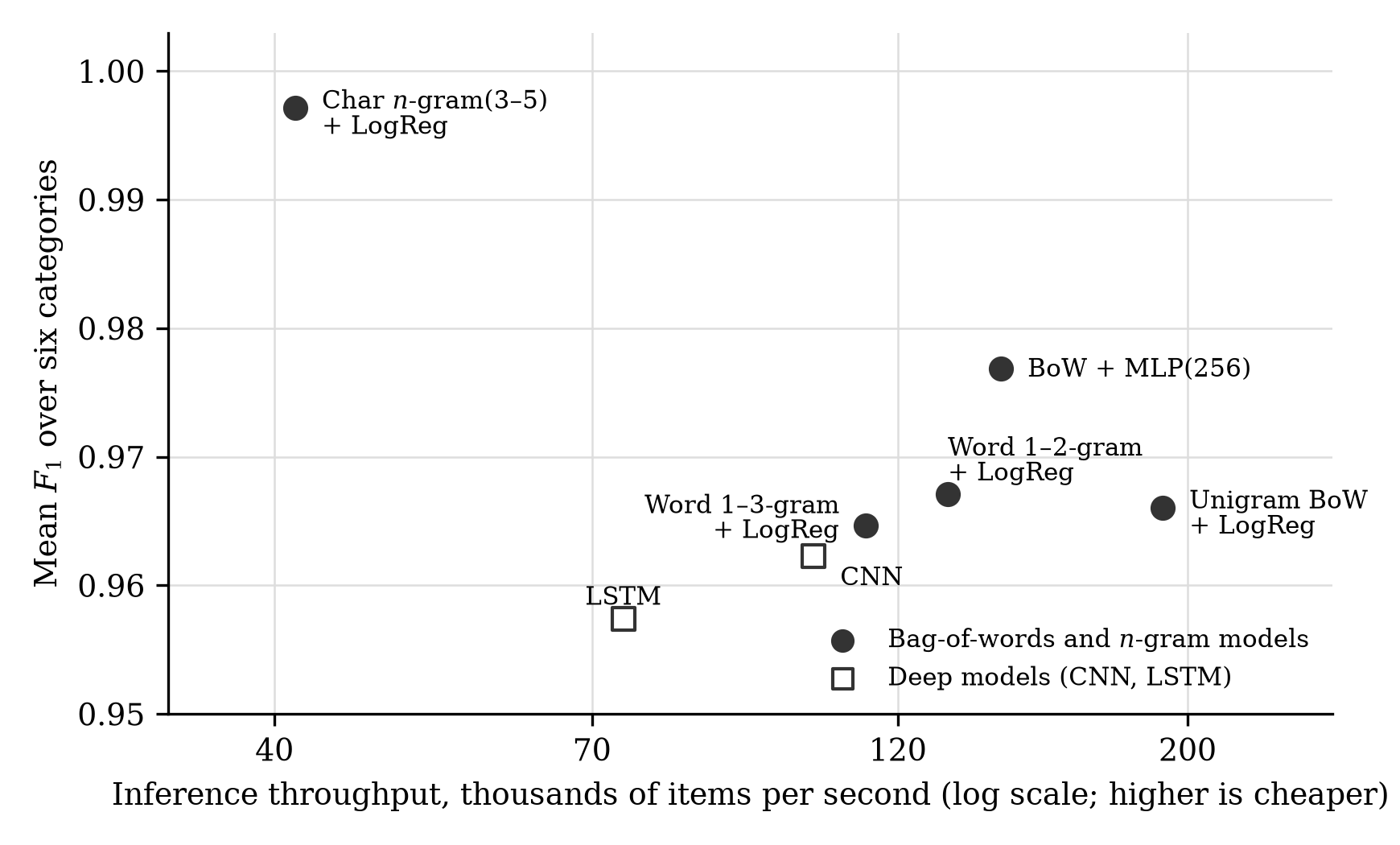}
\end{center}
\caption{\label{fig:cost}Accuracy against measured inference cost, on the
median throughputs of Table~\ref{tab:cost}. A model nearer the top is more
accurate, a model nearer the right is cheaper, and the horizontal axis is
logarithmic. The pretrained encoders of Section~\ref{sec:pretrained} are
excluded because their throughput was measured on a different host and is
comparable with these models only as a ratio (Table~\ref{tab:pretrained}).}
\end{figure}

Describing a model as inexpensive is a claim about measurable quantities, and
this section measures them. Every model was timed under the protocol of
Section~\ref{sec:protocol} on identical hardware: training wall time from raw
strings to a fitted classifier, inference wall time over the test split with
feature extraction included, and the serialized size of everything needed at
inference. Appendix~\ref{app:cost} states the measurement protocol.
Table~\ref{tab:cost} reports medians of time and throughput, and means of
size and parameter count, over the 30 cells, and
Figure~\ref{fig:cost} plots accuracy against inference cost.

The measurement does not support the convenient summary that the most
accurate model is also the cheapest. Character $n$-grams are the slowest of
all seven models at inference, at 41{,}500 items per second against 191{,}300
for unigram bag-of-words, and take 4.0 times as long to train. Character
$3$--$5$-grams generate far more active features per string than word unigrams
do, and on short strings the cost is dominated by feature extraction rather
than by the classifier; the multilayer perceptron, whose serialized model is
59 times larger than the character model's, none the less performs inference
3.5 times faster, for the same reason.

Three models lie on the accuracy--cost frontier, and they are the choice a
practitioner actually faces. Unigram bag-of-words is the cheapest on every
measure, at $F_1 = 0.966$; the multilayer perceptron buys 1.1 $F_1$ points for
about a quarter less throughput; character $n$-grams buy a further 2.0 points
for a further 3.5-fold reduction. The remaining four models are off the
frontier, that is some model is both more accurate and faster to score.

Cost is where the deep models are decisively behind the word-level models.
Trained to the same stopping criterion as everything else, the CNN takes 212
times as long to fit as the cheapest logistic model and the LSTM 258 times,
against a 4.0-fold spread across the four logistic models themselves. Even
the multilayer perceptron, much the largest model here at 2.2\,MiB and
73{,}000 parameters, fits in about half the CNN's time and has a higher mean
$F_1$. Every other model in the comparison therefore has a higher mean $F_1$
and trains faster than either deep model. A model that costs two orders of
magnitude more to fit has to buy something, and on this benchmark neither
does.

Two cautions about the absolute figures. They were measured on the evaluation
split, 330 items, which is small enough that per-call overhead is part of what
is being timed, and it does not penalise the models equally. Repeating the
measurement at 3{,}300, 33{,}000 and 330{,}000 items raises the unigram
model's throughput by roughly half before it plateaus but the character
model's by only 4 to 9\%, because overhead is a larger share of a fast model's
time; the median ratio between them accordingly widens from 4.4 at 330 items
to between 6.1 and 6.8 at the larger batches
(\path{results/throughput_by_batch_summary.csv}). The spread reported in
Table~\ref{tab:cost} is therefore the conservative one. And the timings come
from a shared virtual host, so they are reproducible as ratios rather than as
absolute rates. What the table supports is the comparison between models, not
a capacity estimate for a production deployment, which a per-category binary
measurement over a few hundred strings cannot give. What is established is
that the cost differences between these models are real, spanning a factor of
4.6 at inference and of more than 250 in training. \citet{HarmsSpinder2019}
report training times, and \citet{Oancea2023} the computation time of their
study as a whole, but per-model inference cost had not, to the author's
knowledge, previously been reported for this task.

\begin{table}[hbt]
\caption{\label{tab:cost}Measured cost per model. Training time and throughput
are medians over the 30 category--seed cells, with the standard deviation
across cells in parentheses; size and parameter count are means over the 30
cells. One CPU thread of an Intel Xeon at 2.80\,GHz.
Training time covers fitting the representation and the classifier;
throughput covers feature extraction and scoring; size is the serialized
representation together with the classifier. The parameter count is the
number of fitted coefficients for the four logistic models and the total
number of tensor elements for the convolutional, recurrent and MLP
models, so the two families are not directly comparable on that column.}
\begin{center}\small
\begin{tabular}{lrrrr}
\toprule
Model & Train (s) & Items/s ($10^3$) & Size (KiB) & Parameters \\
\midrule
Unigram bag-of-words + log.\ reg. & $0.012\ (0.004)$ & $191.3\ (33.3)$ & $6.7$ & $283$ \\
Word $1$--$2$-gram + log.\ reg. & $0.017\ (0.003)$ & $131.1\ (19.8)$ & $26.2$ & $1{,}055$ \\
Word $1$--$3$-gram + log.\ reg. & $0.021\ (0.005)$ & $113.5\ (14.8)$ & $53.6$ & $1{,}941$ \\
Char $n$-gram ($3$--$5$) + log.\ reg. & $0.048\ (0.011)$ & $41.5\ (4.6)$ & $38.5$ & $2{,}132$ \\
Unigram bag-of-words + MLP ($256$) & $1.374\ (0.212)$ & $143.9\ (28.7)$ & $2{,}284.8$ & $72{,}705$ \\
CNN (embedding + convolution) & $2.561\ (0.309)$ & $103.4\ (14.1)$ & $66.9$ & $15{,}370$ \\
LSTM (embedding + recurrence) & $3.114\ (0.634)$ & $74.0\ (9.4)$ & $141.1$ & $34{,}250$ \\
\bottomrule
\end{tabular}
\end{center}
\end{table}

\subsection{Pretrained encoders}
\label{sec:pretrained}

Table~\ref{tab:pretrained} sets the three pretrained-encoder configurations
of Section~\ref{sec:pretrained-methods} beside the two models re-timed on the
same host.

Fine-tuned DistilBERT is the only model in the study that comes close to the
character $n$-gram model. Under the matched stopping rule it reaches mean
$F_1 = 0.993$, and 0.991 on test items unseen in training, after a median of
22 epochs. It wins all 30 cells against each of the three word-level logistic
models, the CNN and the LSTM, and 28 of 30 against the multilayer perceptron,
with one tie. It does not overtake the character model: of the 30 cells the
character model wins 18, DistilBERT 3, and 9 are tied, most of them at
$F_1 = 1$, for a mean gap of 0.39 points. The character model thus wins 18
of the 21 non-tied cells, 86\%, which the bands of
Section~\ref{sec:protocol} call a small but consistent difference; the gap is
about one test item per cell, at the resolution limit of the benchmark, so
the defensible reading is that DistilBERT does not pass the character model
and falls slightly but consistently short of it, not that it is clearly
worse. A fixed three epochs gives 0.989; the
matched-stop model beats it in 21 cells to 2 with 7 ties, a reliable
difference by the bands, so, as with the CNN
and LSTM in Appendix~\ref{app:robust}, the short budget understates the
model. Against the character model the three-epoch version takes 2 cells to
23, with 5 ties, and is reliably behind it.

Frozen MiniLM embeddings with a logistic regression reach $F_1 = 0.970$,
which is no better than bag-of-words. They take 17, 16 and 17 of the 30 cells
against the unigram, word $1$--$2$-gram and word $1$--$3$-gram logistic
models, inside the band this paper calls indistinguishable. They lose to the
multilayer perceptron in 22 of 30 cells, a small but consistent difference,
and to the character model in all 30.
Their errors are mostly missed positives, with a mean recall of 0.956 against
a mean precision of 0.984 over the six categories, where fine-tuned DistilBERT
has 0.991 and 0.996 (\path{results/pretrained/pretrained_summary.csv}):
general-purpose sentence semantics, without fine-tuning,
do not separate a target from hard negatives that share its key tokens.

The cost is of a different order. On CPU, with PyTorch limited to one thread
and the tokenizer's own thread pool not limited, fine-tuned DistilBERT
scores 82 items per second, against 38{,}300 for the character model and
193{,}800 for unigram bag-of-words on the same host, which makes it about 460
and 2{,}350 times slower. These ratios are specific to FP32 inference in
PyTorch under that thread limit; GPU inference, an optimized runtime such as ONNX,
or quantization would reduce them, by amounts not measured here. The ratio
between the two re-timed reference models on this host, 5.1, is close to the
4.6 of Table~\ref{tab:cost}, so the comparison carries over between hosts. The serialized DistilBERT classifier
occupies 255\,MiB and has 67 million parameters, against 38\,KiB for the
character model. The MiniLM pipeline is lighter, at 299 items per second and
87\,MiB, but is still 128 times slower than the character model and less
accurate. Fine-tuning took a median of 94 seconds per category and seed on
the GPU, or 13 seconds for three epochs, where the character model fits in
0.06 seconds on one CPU thread of the same Kaggle host.

For the question this paper asks, the pretrained encoder therefore confirms
rather than overturns the ordering. It buys a clear gain over every word-level
and small deep model, and it is the only alternative that approaches the
character model, but it does not pass it and costs about 460 times as much as
the character model at inference, and 2{,}350 times as much as unigram
bag-of-words. Two limits apply. The generator's character-level
corruption favours character features, and a subword tokenizer sees through it
only in part, so on real product names the gap could be smaller or reversed.
And above $F_1 \approx 0.997$ this benchmark has little room left to
discriminate (Appendix~\ref{app:robust}), so the residual gap between the two
best models is small in absolute terms even where it is consistent.

\begin{table}[hbt]
\caption{\label{tab:pretrained}Pretrained encoders against two models
re-timed on the same host. $F_1$ columns are computed as in
Table~\ref{tab:models}; ``Unseen'' is $F_1$ on test items that do not occur in
the training split. Cost columns are medians over the six categories at
seed 0: CPU throughput with PyTorch or the numerical libraries limited to one
thread (the tokenizer's own thread pool was not limited) and tokenization or
feature extraction included, serialized size, and parameter count. Absolute throughput differs
from Table~\ref{tab:cost} because the host differs; ratios are comparable.
Pretrained-encoder $F_1$ values were transcribed from the run log at four
decimal places, so wins and ties against them are counted after rounding to
four places; GPU training is not bit-deterministic.}
\begin{center}\small
\begin{tabular}{lcccrrr}
\toprule
Model & Mean $F_1$ & Seed s.d. & Unseen & Items/s & Size & Parameters \\
\midrule
Char $n$-gram ($3$--$5$) + logistic reg. & $\mathbf{0.997}$ & $0.002$ & $0.996$ & $38{,}272$ & $38$\,KiB & $2{,}084$ \\
Unigram bag-of-words + logistic reg. & $0.966$ & $0.014$ & $0.957$ & $193{,}826$ & $7$\,KiB & $278$ \\
DistilBERT, fine-tuned, matched stop & $0.993$ & $0.005$ & $0.991$ & $82$ & $255$\,MiB & $66{,}955{,}010$ \\
DistilBERT, fine-tuned, 3 epochs & $0.989$ & $0.008$ & $0.985$ & $86$ & $255$\,MiB & $66{,}955{,}010$ \\
MiniLM embeddings + logistic reg. & $0.970$ & $0.011$ & $0.963$ & $299$ & $87$\,MiB & $22{,}713{,}216$ \\
\bottomrule
\end{tabular}
\end{center}
\end{table}

\subsection{Train--test overlap}
\label{sec:overlap}

Because the generator draws from a bounded combinatorial space, 34.1\% of
test items also occur verbatim in the training split. Restricted to test
items unseen in training, the character $n$-gram model scores $F_1 = 0.996$
against $0.997$ on the full split, and under a deduplicate-then-split
protocol it scores $0.997$, so its near-ceiling result is not an artefact of
repetition. The word-level models lose about one $F_1$ point on unseen items.
Appendix~\ref{app:overlap} gives the detail, including the one place where
the ordering of models does move.

\subsection{The rule stage and the recall of the pipeline}
\label{sec:trieresults}

\begin{figure}[hbt]
\begin{center}
\includegraphics[width=0.92\linewidth]{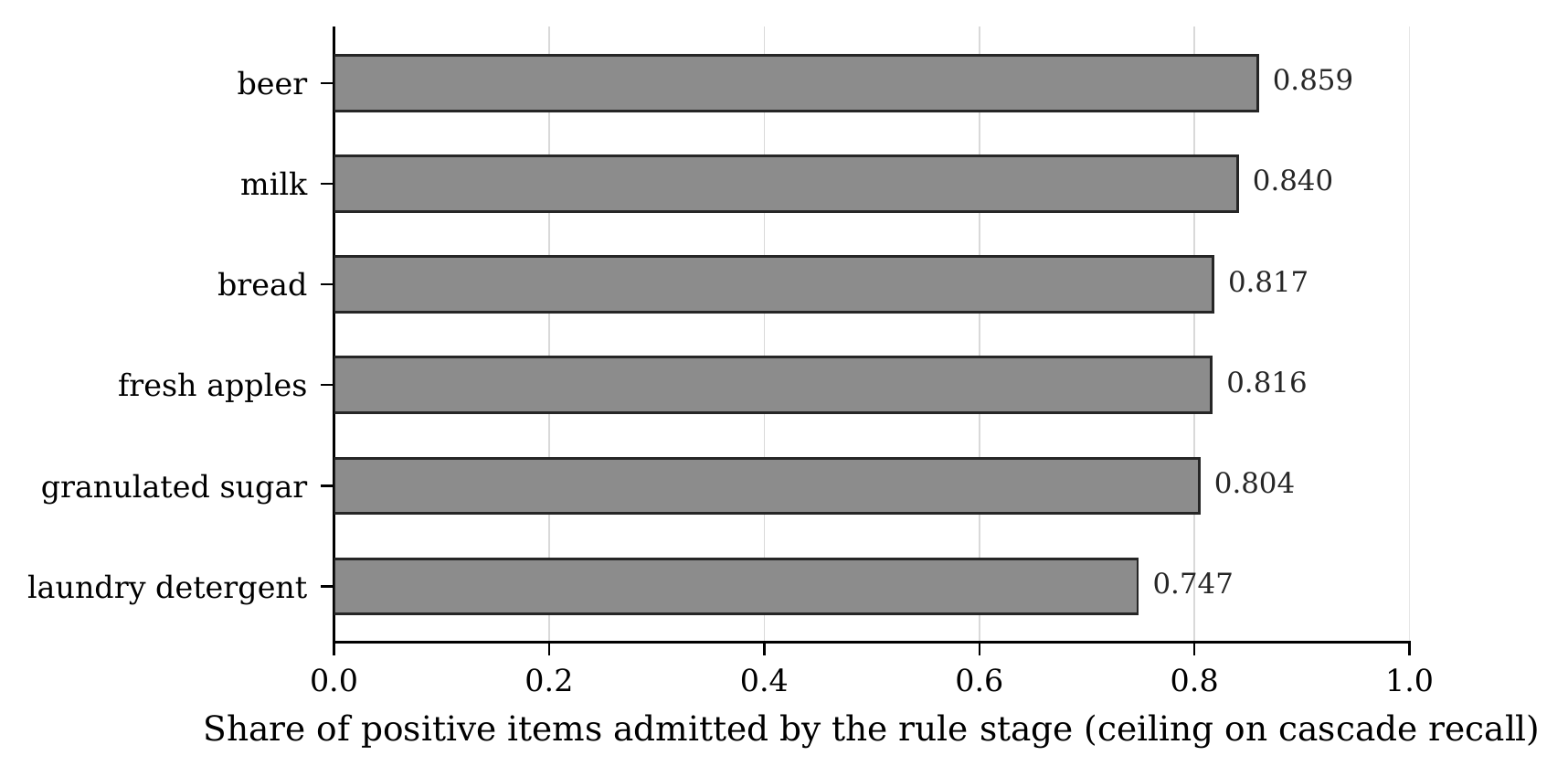}
\end{center}
\caption{\label{fig:trie}Share of positive items admitted by the rule stage,
by category. In a cascade this share is the ceiling on the recall of the
whole pipeline, whatever the confirmation model achieves.}
\end{figure}

Table~\ref{tab:coverage} and Figure~\ref{fig:trie} audit the screening stage.
By construction the positive share of each category corpus is
$700/1650 = 42.4\%$, which bounds the coverage perfect rules could reach;
observed coverage ranges from 31.7\% to 49.9\%. The items screening does not
admit are mostly hard negatives it correctly rejects, between 727 and 950 of
the 950 negatives per category. The positives it loses number 99 to 177 per
category, that is 14.1\% to 25.3\% of positives. Where coverage exceeds the
42.4\% ceiling, in beer, milk and bread, the excess consists of negatives
wrongly admitted, which is why precision on admitted items ranges from 0.729
to 1.000.

Two levels of evaluation must be kept apart. Table~\ref{tab:models} evaluates
the confirmation model in isolation, trained and tested on the full category
corpus. In the cascade of Section~\ref{sec:trie}, items screening does not
admit never reach the confirmation model, so the recall of the whole pipeline
is bounded by the screening recall on positives, 0.747 to 0.859, however
accurate the confirmation model may be. The confirmation stage is needed to
remove the negatives screening wrongly admits; recovering the positives it
misses requires better phrase sets, which is the methodologist's task, or
routing unadmitted items to the model as well, at a higher inference cost.

\begin{table}[hbt]
\caption{\label{tab:coverage}Audit of the rule-screening stage on the
synthetic benchmark, 1650 items per category of which 700 are positive.
Coverage is the share of items admitted to the category and its ceiling under
perfect rules is 0.424; precision is computed on the items the stage admits
and recall over all 700 positives; both are the screening stage's own, not
the confirmation model's.
Because the stage and the generator share phrase sets, these figures are an
optimistic bound.}
\begin{center}
\begin{tabular}{lccccc}
\toprule
Category & Coverage & \multicolumn{2}{c}{Items (count)} & Screening & Screening \\
 & (share) & lost pos. & admitted neg. & precision & recall \\
\midrule
Granulated sugar & $0.366$ & $137$ & $41$ & $0.932$ & $0.804$ \\
Milk & $0.440$ & $112$ & $138$ & $0.810$ & $0.840$ \\
Bread & $0.439$ & $128$ & $152$ & $0.790$ & $0.817$ \\
Beer & $0.499$ & $99$ & $223$ & $0.729$ & $0.859$ \\
Laundry detergent & $0.317$ & $177$ & $0$ & $1.000$ & $0.747$ \\
Fresh apples & $0.346$ & $129$ & $0$ & $1.000$ & $0.816$ \\
\bottomrule
\end{tabular}
\end{center}
\end{table}

\subsection{Data efficiency}
\label{sec:learning}

Table~\ref{tab:learning} shows how little labeled data the confirmation model
needs. With the unigram model, 66 labeled examples, which is 5\% of the
training split, already give $F_1$ between 0.829 and 0.951 depending on
category, rising to 0.951--0.985 on the full split of 1{,}320. Easy
categories are largely solved from a few dozen labels: fresh apples is within
3.1 points of its full-data score at 66 examples. Harder ones gain most from
more data, beer by 13.3 points, bread by 8.7 and granulated sugar by 8.6.

\begin{table}[hbt]
\caption{\label{tab:learning}$F_1$ of the unigram bag-of-words confirmation
model against the number of labeled training examples, averaged over five
split seeds. The test split is held fixed and full size throughout.}
\begin{center}
\begin{tabular}{lrrrrr}
\toprule
& \multicolumn{5}{c}{Labeled training examples} \\
\cmidrule(l){2-6}
Category & $66$ & $132$ & $264$ & $528$ & $1{,}320$ \\
\midrule
Beer & $0.829$ & $0.877$ & $0.913$ & $0.923$ & $0.961$ \\
Bread & $0.871$ & $0.885$ & $0.905$ & $0.925$ & $0.958$ \\
Fresh apples & $0.951$ & $0.954$ & $0.958$ & $0.969$ & $0.982$ \\
Granulated sugar & $0.872$ & $0.906$ & $0.923$ & $0.939$ & $0.958$ \\
Laundry detergent & $0.911$ & $0.936$ & $0.950$ & $0.973$ & $0.985$ \\
Milk & $0.886$ & $0.903$ & $0.921$ & $0.934$ & $0.951$ \\
\bottomrule
\end{tabular}
\end{center}
\end{table}

\subsection{The aggregation rule: recovering labels}
\label{sec:consensus-sim}

\begin{figure}[hbt]
\begin{center}
\includegraphics[width=0.90\linewidth]{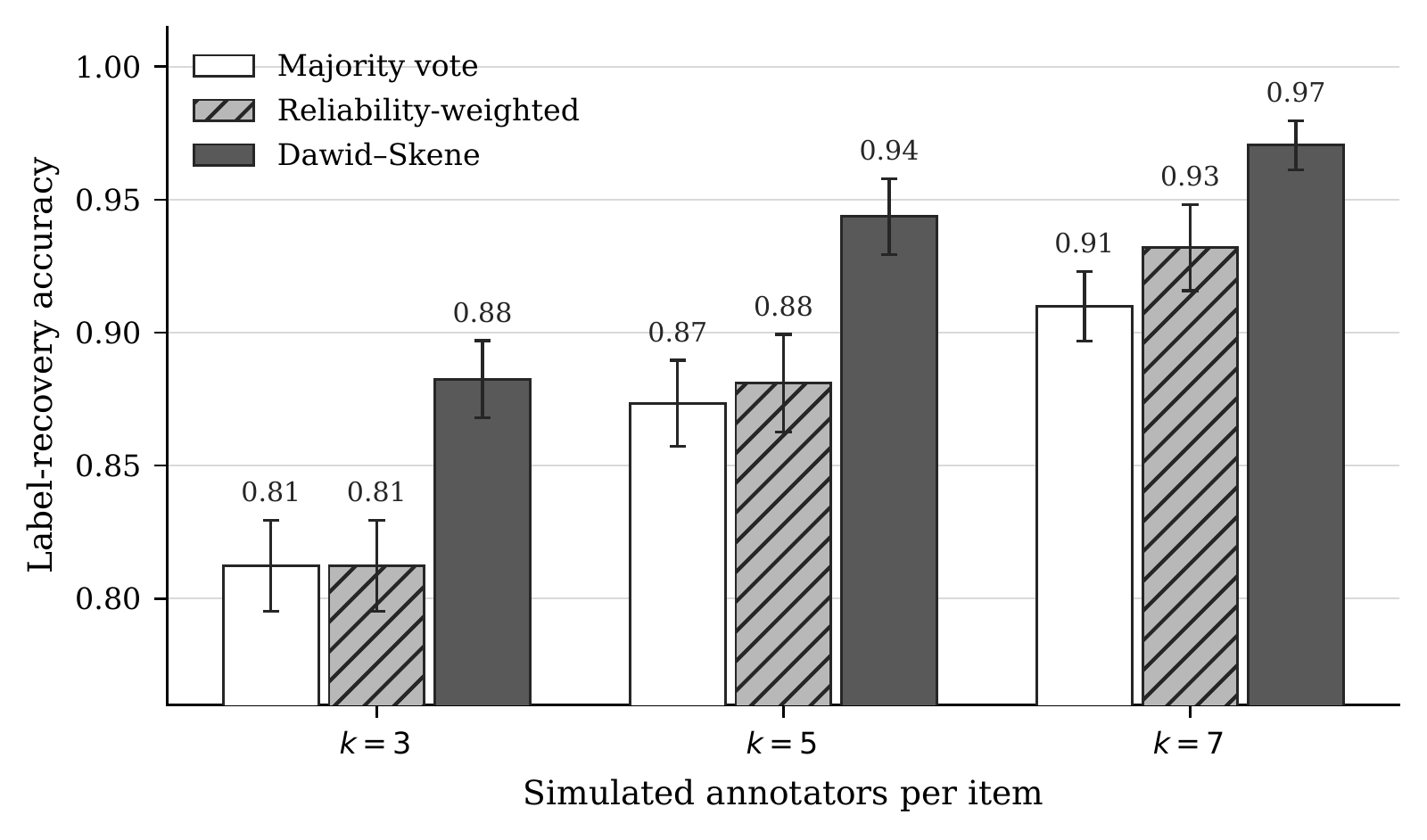}
\end{center}
\caption{\label{fig:consensus}Label-recovery accuracy of three aggregation
rules by number of votes per item, over 60 runs of 1{,}500 items with 12
simulated annotators. Whiskers are $\pm 1$ standard deviation across runs.
The vertical axis is truncated below 0.76 and does not start at zero.}
\end{figure}

The labeling protocol of Section~\ref{sec:labeling} is evaluated here as one
design choice among three. The annotators are simulated and no human
annotation was collected for this study. The simulation generates 1{,}500
items with balanced binary ground truth and 12 annotators from a mixed
population: five expert, with accuracy drawn uniformly from 0.85 to 0.97,
five mediocre, from 0.58 to 0.72, and two adversarial, from 0.35 to 0.50.
Each item
receives $k \in \{3,5,7\}$ independent votes from distinct annotators, and
each vote reproduces the true label with the annotator's accuracy. Three
aggregators recover the true label: plain majority vote; the
reliability-weighted vote of Equations~\ref{eq:consensus}
and~\ref{eq:update}; and Dawid--Skene EM \citep{DawidSkene1979}.
Figure~\ref{fig:consensus} reports label-recovery accuracy over 60 runs.

Dawid--Skene dominates, by six to seven accuracy points over majority vote
and by 3.9 to 7.0 over the reliability-weighted rule, because it estimates
each annotator's full error profile rather than a single scalar. The
reliability-weighted rule barely improves on plain majority: the two are
identical at $k = 3$, differ by under one point at $k = 5$ and by about two
at $k = 7$. The reason follows from Equation~\ref{eq:update}, and the final
weights confirm it (\path{results/consensus_final_weights.csv}). With an
additive increment and a fixed cap, an annotator who agrees with the
consensus more often than not drifts to $r_{\max}$: the median final weight is
2.0 for expert and for mediocre annotators at every $k$. At $k = 3$ even the
adversarial annotators end near the cap, with a median of 1.9, plausibly
because each vote is a third of the consensus it is scored against and that
consensus is itself wrong on a fifth of the items, so the weights are near-uniform and the weighted vote collapses to
majority. Only as $k$ grows does the consensus become accurate enough to pull
adversarial annotators down, to a median of 1.4 at $k = 5$ and 0.4 at
$k = 7$, and that is where the weighted rule's small gain over majority
appears.

Appendix~\ref{app:annotators} shows that Dawid--Skene's advantage persists,
at 3.0 to 3.5 points, when there are no adversarial annotators, and widens to
8.7 to 10.8 points over majority vote when annotators miss positives more
often than they raise false positives.

This is a negative result about a rule adopted in production. A transparent
scalar weight per
assessor is defensible as an operational and audit requirement; it is not
defensible as a statistical optimum. Where label accuracy is what matters, a
latent-ability estimator, or at minimum a multiplicative log-odds weight that
does not saturate, is the better choice.

\subsection{The aggregation rule: training classifiers}
\label{sec:agg-train}

Recovering labels is not the coding task. To test whether the aggregation
rule matters downstream, the training labels of every category and split seed
were replaced with simulated crowd votes from the same annotator mixture,
aggregated by each rule, and the confirmation models were trained on the
result and scored on clean test labels. Table~\ref{tab:aggtrain} reports the
outcome.

For the character $n$-gram model the choice matters: Dawid--Skene yields
classifiers 1.2 to 1.8 $F_1$ points better than majority vote at every $k$,
winning 26 of the 30 paired cells at $k = 3$, a small but consistent
difference by the bands of Section~\ref{sec:protocol}, and 30 and 29 at $k = 5$
and $k = 7$, a reliable one. The reliability-weighted rule is
indistinguishable from majority vote at $k = 3$, where 29 of the 30 cells are
tied, and at $k = 5$, where it takes 9 cells to 5; only at $k = 7$ is it
ahead, by 0.6 points, taking 18 cells to 4 with 8 ties, a small but consistent
difference, and there Dawid--Skene is still ahead of it by a further 0.6
points, taking 22 cells to 6 with 2 ties
(\path{results/aggregation_to_training_pairs.csv}). For the unigram model all
differences stay below one $F_1$ point, because that model's ceiling is lower and label noise is not its
binding constraint. Label quality therefore propagates to the classifier, and
the aggregation rule is a design decision with measurable, if bounded,
consequences.

\begin{table}[hbt]
\caption{\label{tab:aggtrain}Test $F_1$ of confirmation models trained on
aggregated simulated crowd labels, averaged over six categories and five
split seeds. Test labels are clean throughout; only the training labels are
replaced.}
\begin{center}
\begin{tabular}{llcccc}
\toprule
Model & $k$ & Gold labels & Majority & Reliability-weighted & Dawid--Skene \\
\midrule
Char $n$-gram & 3 & $0.997$ & $0.956$ & $0.956$ & $0.973$ \\
Char $n$-gram & 5 & $0.997$ & $0.970$ & $0.971$ & $0.988$ \\
Char $n$-gram & 7 & $0.997$ & $0.979$ & $0.986$ & $0.992$ \\
Unigram bag-of-words & 3 & $0.966$ & $0.952$ & $0.952$ & $0.952$ \\
Unigram bag-of-words & 5 & $0.966$ & $0.956$ & $0.958$ & $0.964$ \\
Unigram bag-of-words & 7 & $0.966$ & $0.958$ & $0.959$ & $0.965$ \\
\bottomrule
\end{tabular}
\end{center}
\end{table}

\section{Discussion}
\label{sec:discussion}

\subsection{What the benchmark does and does not show}

All quantitative results here are synthetic. They establish the relative
ordering of models under a stated protocol, not absolute accuracy in
production. Five limits bound every claim above.

The generator applies character-level corruption, which creates conditions
favourable to character $n$-gram features, while the CNN and LSTM see
corrupted tokens only as unknown symbols. A different noise structure, or
character-aware neural models \citep{Zhang2015charcnn}, could reorder the
models, and in part it does: when the corpus is regenerated with word-level
noise in place of character-level corruption (Appendix~\ref{app:noise}), the
character model stays first but its lead over the word-level models falls
from about 3 points to about 0.5, and the multilayer perceptron no longer
leads the word-level logistic models, while without any noise every model
scores $F_1 = 1$. The correct reading of Section~\ref{sec:models} is
therefore that a linear model with sparse character features is sufficient
on this benchmark, not that character features are generally optimal for
product coding.

The generator also plants a shortcut. Positive items carry a grade block with
probability 0.7 by construction, and between 55\% and 63\% of positive strings
carry an intact grade word outside their key phrase (73\% to 83\% counting
grade words that are also key-phrase tokens, such as \emph{white sugar} or
\emph{whole milk}), against none of the negatives. Removing the grade block
from both training and test strings (Appendix~\ref{app:grade}) costs the
word-level logistic models 1.0 to 1.35 $F_1$ points and the multilayer
perceptron 1.0, but the character model only 0.2, and leaves the character
model first and the perceptron second. The cue inflates the absolute scores of the word-level
models more than it drives the ordering, but a real corpus would have its own
cues, and the benchmark is English-language, whereas the deployment it stands
in for is not.

The rule stage and the generator share their phrase sets, so screening is
evaluated on data produced from its own vocabulary. Its coverage and
precision in Table~\ref{tab:coverage} are an optimistic bound, and the only
claim that survives this circularity is the structural one: in a cascade,
screening recall caps pipeline recall.

The five seeds within a category are repeated splits of one corpus, so the
paired comparisons of Section~\ref{sec:protocol} do not rest on independent
replications \citep{Dietterich1998, NadeauBengio2003}. Nor, strictly, do the categories: all six come from one
generator applying one noise process, so they share more structure than six
product groups drawn from a real feed would. Consistency across the 30 cells
is the meaningful evidence; the $p$-values are not.

The benchmark covers six categories and a binary decision per category.
Multi-class assignment across a full COICOP hierarchy, including items
admitted by more than one category, is outside its scope, as is the
behaviour of any of these models under concept drift \citep{Meertens2022}.

\subsection{The cost result}

Section~\ref{sec:cost} supplies a measurement the literature has mostly left
out: \citet{HarmsSpinder2019} report training time, 43 seconds against more
than nine hours, but per-model inference cost had not, to the author's
knowledge, previously been reported for this task, and the result is worth isolating. The shorthand that cheap models win bundles two
distinct statements: that lightweight models are accurate enough, and that
the accurate ones are cheap. On this benchmark the first holds and the second
does not. The most accurate model is the slowest
at inference of the seven, because feature extraction rather than
classification dominates the cost on short strings, and the model that is
cheapest on every measure is three $F_1$ points behind it. A practitioner
choosing between them is making a real trade.

The finding about deep models needed correcting too, in the other direction.
The version of this benchmark in an earlier version (v3) of the preprint
\citep{Beskorovainyi2026arxiv} trained them for a fixed eight epochs while every other model ran to convergence, and
concluded they were the least accurate models in every category. Under a
matched stopping criterion that conclusion does not hold: they are not
distinguishable from the word-level logistic models on accuracy. What does
hold is that neither has a higher mean $F_1$ than any non-deep model and that
they cost 212 and 258 times as much to fit as the cheapest model in the
comparison, which is the weaker claim the evidence supports. The direction is still consistent with \citet{HarmsSpinder2019} on
real data, and with the explanation they offer, that short non-cohesive
product descriptions give order-sensitive architectures nothing to exploit;
the strength of the effect was an artefact of the budget, and
Appendix~\ref{app:robust} reports it.
The qualification in Section~\ref{sec:models} still applies: these are small
and lightly tuned networks. A fine-tuned pretrained encoder \citep{Devlin2019}
does better: Section~\ref{sec:pretrained} shows DistilBERT, fine-tuned end to
end, reaching within 0.4 $F_1$ points of the character model, but at several
hundred times its inference cost, while frozen MiniLM embeddings do not beat
the multilayer perceptron. Larger encoders, and the retrieval-augmented language models
now being evaluated for this task \citep{Toth2025}, were not tested.

\subsection{Design choices and their costs}

A single global classifier over a large, shifting label set is brittle when
the basket changes and when per-category data are imbalanced. Screening
encodes methodologist knowledge transparently and removes the easy,
high-volume cases cheaply; per-category binary confirmation then absorbs the
hard residual and can be retrained in isolation. The cost of this design is
operational complexity, since there are many models to monitor and many
category definitions to maintain, and that burden is real enough that the
quality frameworks developed for intermediate machine-learning outputs
\citep{Yung2022, Molladavoudi2023} apply directly to it. A modern alternative
would replace the per-category classifiers with a single pretrained encoder
fine-tuned for hierarchical classification \citep{Beresewicz2024}; the
auditability and per-category isolation of the present design are valued in a
regulated setting, but that is an argument for weighing the alternative, not
for dismissing it.

Folding the confirmation model into the consensus as an additional voter, as
Section~\ref{sec:weights} describes, risks a feedback loop in which the model
reinforces its own errors. The mitigations are to cap the model's effective
weight below $r_{\max}$ and to route a fixed fraction of items to human-only
labeling. Neither is evaluated here, and the risk should be regarded as open.

\subsection{Coverage and representativeness}

Transaction and receipt sources over-represent outlets and product types that
are fully digitized, and the join and validation steps remove records, so
coverage is not uniform. Coverage statistics and comparison against
field-collected quotes are essential before such quotes substitute for,
rather than augment, traditional collection. This paper does not supply
either.

\subsection{The priority next step}

The single most valuable extension is an evaluation on a public, real
product-name corpus under the same matched protocol, with the same dispersion
reporting, paired comparison, overlap audit and cost measurement.
\citet{Montbroussous2024} give the sharpest reason: on their data a model
scoring 97.4\% on pre-labelled test items coded only 41.4\% of expenditure
correctly on genuinely unlabelled data. Any near-ceiling figure on a
controlled corpus, including every figure in Section~\ref{sec:models}, should
be read in that light. The released package is structured so that a real
corpus is a drop-in addition, and this is flagged as the next step rather
than claimed here.

\section{Conclusions}
\label{sec:conclusion}

That lightweight text classifiers suffice for coding short retail product
names to consumer-price categories is not a new finding; statistical offices
have reported it on real data. What this paper adds is the evaluation that
makes the finding checkable: one matched protocol across seven models trained
from scratch and two pretrained encoders in three configurations,
dispersion across seeds, paired comparison, an audit of train--test overlap,
a separate audit of the rule stage that precedes the classifier, and a
measurement of what each model costs.

Three things are established within the limits of a synthetic benchmark. A
character $n$-gram logistic regression is the most accurate model in every
category and wins every paired comparison against the other models trained
from scratch, by a margin that depends in part on the generator's
character-level noise, while neither the small CNN nor the LSTM examined here has a
higher mean $F_1$ than any non-deep model; they are not reliably worse than
the word-level models once their training budget is matched, and cost 212 and
258 times as much to fit for that parity. A fine-tuned pretrained encoder comes within half a point of the
character model but does not pass it, at several hundred times its inference
cost. The most accurate model is not
the cheapest: character features cost a 4.6-fold reduction in inference
throughput relative to unigram bag-of-words, which is the trade a
practitioner actually faces and which the unmeasured version of the claim
leaves out. And the aggregation rule used to produce training labels matters
downstream for the character model, with an additive reliability weight performing close to majority
vote because it saturates, and a latent-ability estimator better than both.

Two things remain uncertain and are the gate before transaction data can move
from augmenting to partially replacing manual price collection: how much of
the basket such data actually cover, and how closely receipt-derived quotes
track field-collected ones. Neither is answered here. What should be tested
next is this same protocol on a public, real corpus of product names, where
the ordering reported above can be confirmed or overturned on evidence anyone
can reproduce.

\section*{Acknowledgements}

The author thanks the reviewers of an earlier version of this manuscript,
whose insistence on measuring computational cost rather than asserting it
produced the result reported in Section~\ref{sec:cost}.

\section*{Data and code availability}

All data, models and scripts are available on GitHub
(\url{https://github.com/beskvladimir-create/cpi-coding-repro}) and the exact
release from which every accuracy figure outside the pretrained-encoder
sections is reproduced, release 2.1.1, is archived on Zenodo (DOI:
\texttt{10.5281/zenodo.23445113}). The script for the pretrained-encoder
baselines of Section~\ref{sec:pretrained}, the notebook that ran it, the cost
file it wrote and the per-run and per-category values transcribed from the
notebook's log and output preview are provided as Supplemental File~1, which
marks each file as written or transcribed; the script records the exact model
revisions used.
The benchmark is fully synthetic and contains no proprietary, personal or
production data.

\section*{Funding}

This research received no external funding.

\section*{Competing interests}

The author operates Besk Tech, which develops software products based on
language models. The architecture described in this paper was developed by the
author in previous employment; the author has no current commercial interest
in it, and this work received no funding.

\appendix

\section{The synthetic generator}
\label{app:gen}

Each of the six categories is defined by a set of key phrases that act as
positive product cores, a set of stop phrases that act as exclusions, and
lists of brands, grades, units, fillers and hard-negative cores. The same key
and stop phrase sets are consumed by the rule stage of
Section~\ref{sec:trie}, so the stage and the data share one source of truth;
Section~\ref{sec:discussion} states what this circularity does and does not
permit.

A positive string is assembled in fixed block order as grade, brand, key
phrase, unit, filler, where the grade block is present with probability 0.7,
the brand block with 0.6, the unit block with 0.7 and the filler block with
0.4. A negative string is assembled the same way from a hard-negative core,
without a grade block. Cores are drawn uniformly.

Each assembled string receives token-level noise with probability 0.30.
Within a noised string, every token longer than three characters is altered
with probability 0.5 by one of three transformations chosen uniformly:
truncation by one or two trailing characters, deletion of all vowels, applied
only where at least three consonants remain, or transposition of two adjacent
characters. A noised string additionally has two adjacent tokens concatenated
with probability 0.3.

Per category the generator draws 700 positives and 950 hard negatives from
corpus seed 0 and shuffles them. Between 73.0\% and 78.1\% of the strings in
each corpus are distinct; Appendix~\ref{app:overlap} quantifies the
consequence for train--test overlap. A manifest file records the seed, the
noise rate and the phrase sets for every category.

\section{Train--test overlap}
\label{app:overlap}

Because the generator draws from a bounded combinatorial space, a test string
can recur verbatim in training. Averaged over the six categories and five
split seeds, 34.1\% of test items also occur in the training split, made up,
by category, of 19.3\% to 26.4\% of positive test items and 38.2\% to 46.3\% of negative
ones. No string occurs with conflicting labels.

Two stricter protocols were therefore applied, to the four logistic models;
the multilayer perceptron and the two deep models were not re-evaluated, so
no statement
here covers them. Table~\ref{tab:overlap} reports the result.

\begin{table}[hbt]
\caption{\label{tab:overlap}Mean $F_1$ of the four logistic models on the
full test split, on test items unseen in training, and under a
deduplicate-then-split protocol, averaged over six categories and five seeds.
Four decimal places are given because the margins between the three
word-level models are smaller than three places resolve.}
\begin{center}
\begin{tabular}{lccc}
\toprule
Model & Full split & Unseen only & Deduplicated \\
\midrule
Char $n$-gram ($3$--$5$) & $0.9972$ & $0.9963$ & $0.9970$ \\
Word $1$--$2$-gram & $0.9671$ & $0.9580$ & $0.9634$ \\
Unigram bag-of-words & $0.9661$ & $0.9568$ & $0.9616$ \\
Word $1$--$3$-gram & $0.9647$ & $0.9548$ & $0.9625$ \\
\bottomrule
\end{tabular}
\end{center}
\end{table}

The character $n$-gram model keeps its position under both protocols, losing
0.1 $F_1$ points on unseen items, so its near-ceiling score is not an
artefact of repetition. The word-level models lose about one point. Their
ordering among themselves is not stable: under deduplication the word
$1$--$3$-gram model overtakes unigram bag-of-words, by 0.1 $F_1$ points.
Given that the three word-level sets lie within 0.3 points of one another on
the full split, that reversal should be read as confirming they are
interchangeable on this benchmark rather than as a re-ranking.

\section{Robustness of the model comparison}
\label{app:robust}

Six checks test whether the ordering in Section~\ref{sec:models}, and the
aggregation result of Section~\ref{sec:consensus-sim}, depend on choices made
for convenience or on properties of the generator. They are produced by
\path{src/robustness_checks.py}, \path{src/aggregation_passes.py},
\path{src/consensus_sensitivity.py}, \path{src/grade_token_ablation.py} and
\path{src/noise_sensitivity.py} in the released package, with outputs under
\path{results/}. The last three regenerate corpora or simulations in memory
and never overwrite the released data.

\subsection{Training budget}

The comparison is sensitive to how long the deep models are trained, which is
why Section~\ref{sec:protocol} gives every model a convergence criterion
rather than a fixed number of passes. Table~\ref{tab:epochs} shows what a
fixed budget, or validation-based early stopping, would have produced
(\path{results/robustness_training_budget.csv}).

\begin{table}[hbt]
\caption{\label{tab:epochs}Mean $F_1$ of the two deep models under four
training budgets, over the 30 category--seed cells. The first column is the
fixed budget used in an earlier version of this benchmark; the third is the
matched rule used throughout this paper; the fourth holds out a stratified
20\% of the training split, trains for 40 epochs on the remaining 80\%, and
keeps the epoch with the best validation $F_1$.}
\begin{center}
\begin{tabular}{lcccc}
\toprule
Model & 8 epochs, & 40 epochs, & Matched & Validation \\
 & fixed & fixed & rule & early stop \\
\midrule
CNN & $0.946$ & $0.962$ & $0.962$ & $0.955$ \\
LSTM & $0.942$ & $0.957$ & $0.957$ & $0.949$ \\
\bottomrule
\end{tabular}
\end{center}
\end{table}

A fixed eight epochs understates the CNN by 1.6 $F_1$ points and the LSTM by
1.5, which is more than the 0.3 points separating the three word-level
logistic models from one another. Under that budget both deep models sit below
every other model in every category, and the comparison invites the conclusion
that order-sensitive architectures are worse at this task. Under a matched
criterion they are not distinguishable from the word-level models, and the
conclusion that survives is the narrower one stated in
Section~\ref{sec:models}. Validation-based early stopping scores lower, 0.955
for the CNN and 0.949 for the LSTM, with the best validation epoch at a median
of 21.5 and 20 respectively; a plausible reason is that the validation split
costs a fifth of the training examples on a corpus this small.

\subsection{Decision threshold}

All $F_1$ figures use the default threshold of $0.5$. Under average precision,
which integrates over every threshold, the mean over the 30 cells is 0.99997
for the character $n$-gram model, 0.99799 for the multilayer perceptron,
0.99676 for word $1$--$2$-grams, 0.99650 for word $1$--$3$-grams, 0.99642 for
unigram bag-of-words, 0.99537 for the CNN and 0.99154 for the LSTM
(\path{results/robustness_average_precision_summary.csv}). The ordering is
that of $F_1$ except that unigram bag-of-words and the word $1$--$3$-gram model
exchange fourth and fifth place; Section~\ref{sec:models} already reports
those two as indistinguishable. The character model's average precision is
exactly 1 on 8 of the 30 splits and no lower than 0.99980 on any. This is
worth stating plainly as a limit of the benchmark rather than as a result: the
character model ranks the two classes almost perfectly, and above it the
corpus has little room left to discriminate between methods.

\subsection{Batch against online aggregation}

Section~\ref{sec:consensus-sim} compares an online rule, which updates weights
in a single pass over the items, against Dawid--Skene, which sees every item
at once and iterates. That difference in information could by itself explain
the gap. It does not (\path{results/aggregation_passes.csv}). Giving the
reliability-weighted rule two, five or twenty passes, carrying the weights
over between them, changes its accuracy by at most 0.11 points, at $k = 7$
with two passes, and by at most 0.04 points at $k = 3$ and $k = 5$. The gap to
Dawid--Skene is 4.05 to 7.37 points with one pass and 3.94 to 7.38 points
across all pass counts. This simulation draws ground truth independently for
each item rather than as an exactly balanced set, so its figures differ
slightly from those of Figure~\ref{fig:consensus}. The saturation mechanism of
Section~\ref{sec:weights}, not the number of passes, is what bounds the rule.

\subsection{Annotator population}
\label{app:annotators}

The simulation of Section~\ref{sec:consensus-sim} fixes one annotator
population, with two adversarial annotators among twelve and errors that do
not depend on the true class. Two departures were run with everything else
unchanged, including the seeds (\path{results/consensus_sensitivity.csv}).
In the first the two adversarial annotators are replaced by mediocre ones, so
the population is five expert and seven mediocre annotators. In the second
the released population is kept, but an annotator of accuracy $a$ recognises a
true positive with probability $a - 0.10$ and a true negative with probability
$a + 0.10$, so positives are missed more often than false positives are
raised while accuracy over the balanced item set stays $a$.
Table~\ref{tab:annot} reports the result.

\begin{table}[hbt]
\caption{\label{tab:annot}Label-recovery accuracy of the three aggregation
rules under the released annotator population and two alternatives, mean
over 60 runs of 1{,}500 items.}
\begin{center}\small
\begin{tabular}{llccc}
\toprule
Population & $k$ & Majority & Reliability-weighted & Dawid--Skene \\
\midrule
Released (Section~\ref{sec:consensus-sim}) & 3 & $0.812$ & $0.812$ & $0.882$ \\
 & 5 & $0.873$ & $0.881$ & $0.944$ \\
 & 7 & $0.910$ & $0.932$ & $0.970$ \\
No adversarial annotators & 3 & $0.854$ & $0.854$ & $0.887$ \\
 & 5 & $0.911$ & $0.911$ & $0.946$ \\
 & 7 & $0.942$ & $0.942$ & $0.972$ \\
Class-dependent errors & 3 & $0.794$ & $0.794$ & $0.881$ \\
 & 5 & $0.841$ & $0.844$ & $0.950$ \\
 & 7 & $0.868$ & $0.884$ & $0.975$ \\
\bottomrule
\end{tabular}
\end{center}
\end{table}

Dawid--Skene's advantage persists in both. Without adversarial annotators it
narrows to 3.0 to 3.5 points over both other rules, and the
reliability-weighted rule becomes indistinguishable from majority vote at
every $k$, differing by less than 0.01 points, because with no annotator
below chance every weight saturates at $r_{\max}$
(\path{results/consensus_sensitivity_weights.csv}). With class-dependent
errors the advantage widens to 8.7 to 10.8 points over majority vote and 8.7
to 10.6 over the weighted rule, since Dawid--Skene estimates separate error
rates for the two classes and the two vote-counting rules cannot; the
weighted rule's gain over majority vote stays at 0.25 points or less at
$k = 3$ and $k = 5$ and is 1.6 points at $k = 7$.

\subsection{Grade-token shortcut}
\label{app:grade}

The generator of Appendix~\ref{app:gen} adds a grade block to positive strings
with probability 0.7 and never to negative ones, so a grade word is a cue that
a model can use without reading the product core. Depending on category,
between 55\% and 63\% of positive strings carry an intact grade word outside
their key phrase, and between 73\% and 83\% carry one anywhere, counting grade
words that are also key-phrase tokens, such as \emph{white} in \emph{white
sugar} or \emph{whole} in \emph{whole milk}; no negative string carries one
(\path{results/grade_word_shares.csv}).

To measure how much the models rely on the cue, the corpus was regenerated in
memory from the same seed and the same random draws, and the grade block was
then removed wherever it had ended up after noise: as a token of its own, in
corrupted form, or as the leading part of a concatenated token. Key-phrase
tokens were never removed, so \emph{pale ale} and \emph{whole milk} stay
intact, and the two corpora differ in the grade block alone; the regenerated
original is checked to be identical to the released one. The four logistic models and the
multilayer perceptron were then re-run under the protocol of
Section~\ref{sec:protocol}; the CNN and LSTM were not re-run.
Table~\ref{tab:grade} reports the result
(\path{results/grade_token_ablation_summary.csv}, from
\path{src/grade_token_ablation.py}).

\begin{table}[hbt]
\caption{\label{tab:grade}Mean $F_1$ over the 30 category--seed cells on the
released corpus and on the same corpus with the grade block removed from
training and test strings. The last column counts the cells in which removal
lowers, raises or leaves unchanged the model's $F_1$.}
\begin{center}\small
\begin{tabular}{lcccc}
\toprule
Model & Original & Grade block & Change & Cells lower / \\
 & & removed & (points) & higher / same \\
\midrule
Char $n$-gram ($3$--$5$) & $0.9972$ & $0.9952$ & $-0.20$ & $10 / 4 / 16$ \\
Unigram bag-of-words + MLP ($256$) & $0.9769$ & $0.9670$ & $-0.99$ & $27 / 3 / 0$ \\
Word $1$--$2$-gram & $0.9671$ & $0.9536$ & $-1.35$ & $25 / 4 / 1$ \\
Unigram bag-of-words & $0.9661$ & $0.9559$ & $-1.02$ & $23 / 7 / 0$ \\
Word $1$--$3$-gram & $0.9647$ & $0.9517$ & $-1.30$ & $25 / 5 / 0$ \\
\bottomrule
\end{tabular}
\end{center}
\end{table}

The character model barely depends on the cue, and its lead over the other
models widens. The multilayer perceptron stays second. The three word-level
models lose 1.0 to 1.35 points and remain within 0.5 points of one another,
unigram bag-of-words now narrowly ahead of word $1$--$2$-grams, two models
that Section~\ref{sec:models} already finds indistinguishable. The ordering
of Table~\ref{tab:models} among these five models is therefore unchanged
except within the word-level group.

\subsection{Generator noise}
\label{app:noise}

The generator's character-level corruption favours character features
(Section~\ref{sec:discussion}). To see how far the ordering depends on it, the
six corpora were regenerated in memory from the same seed and the same random
draws, changing only the noise process: no noise at all; the released
character-level noise applied to 30\% of strings, which reproduces the
released corpus string for string; the same noise applied to 60\% of strings;
and word-level noise only, at the released rate. The released generator has
one word-level operation, the concatenation of two adjacent tokens, and three
character-level ones; the word-level variant keeps the concatenation and
replaces the three character-level operations, with the same probabilities,
by deletion or duplication of a whole token, so that every surviving token is
correctly spelt. All seven models were re-run under the protocol of
Section~\ref{sec:protocol}; on the released corpus they reproduce the
published per-cell $F_1$. Table~\ref{tab:noise} reports the result
(\path{results/noise_sensitivity_summary.csv}, with paired counts and bands in
\path{results/noise_sensitivity_pairs.csv}).

\begin{table}[hbt]
\caption{\label{tab:noise}Mean $F_1$ over the 30 category--seed cells when the
corpus is regenerated with a different noise process. ``Char.'' is the
released character-level noise, applied to the stated share of strings; the
0.3 column is the released corpus. ``Word'' replaces the character-level
operations by token deletion or duplication. The last row is the share of
distinct strings in a corpus, averaged over categories.}
\begin{center}\small
\begin{tabular}{lcccc}
\toprule
Model & No noise & Char., 0.3 & Char., 0.6 & Word, 0.3 \\
\midrule
Char $n$-gram ($3$--$5$) + logistic reg. & $1.0000$ & $0.9972$ & $0.9977$ & $0.9732$ \\
Unigram bag-of-words + MLP ($256$) & $1.0000$ & $0.9769$ & $0.9678$ & $0.9653$ \\
Word $1$--$2$-gram + logistic reg. & $1.0000$ & $0.9671$ & $0.9531$ & $0.9676$ \\
Unigram bag-of-words + logistic reg. & $1.0000$ & $0.9661$ & $0.9515$ & $0.9675$ \\
Word $1$--$3$-gram + logistic reg. & $1.0000$ & $0.9647$ & $0.9523$ & $0.9683$ \\
CNN (embedding + convolution) & $1.0000$ & $0.9623$ & $0.9480$ & $0.9629$ \\
LSTM (embedding + recurrence) & $1.0000$ & $0.9574$ & $0.9415$ & $0.9629$ \\
\midrule
Distinct strings (share) & $0.605$ & $0.756$ & $0.875$ & $0.728$ \\
\bottomrule
\end{tabular}
\end{center}
\end{table}

Without noise every model scores $F_1 = 1$ in every cell: the hard negatives
are separable from their key phrases alone, so the benchmark discriminates
between models only through its noise. With the noise rate doubled, the
ordering of Table~\ref{tab:models} holds and the gaps widen: the character
model is unaffected, at 0.998, and wins all 30 cells against every other
model, and the multilayer perceptron wins 28 to 30 of the 30 cells against
each word-level model, which lose 1.2 to 1.5 points. Under word-level noise
the ordering does not survive in full. The character model is still first,
but its lead shrinks to 0.5 to 0.6 points over the word-level models, a
reliable or small but consistent difference (it takes 26 or 27 of their 30
cells), and to 0.8 points over the multilayer perceptron (25 cells to 4). The
multilayer perceptron loses its second place: the three word-level logistic
models are 0.2 to 0.3 points ahead of it, indistinguishable by the bands, and
the CNN and LSTM are within 0.25 points of it. The advantage of character
features is therefore partly, though not wholly, a product of the generator's
character-level corruption, and the advantage of the perceptron over
word-level logistic regression appears to depend on it entirely.

\section{Cost measurement}
\label{app:cost}

Timings in Table~\ref{tab:cost} were taken under the protocol of
Section~\ref{sec:protocol}, one measurement per category and seed, giving 30
measurements per model. Training time is wall-clock time from raw training
strings to a fitted classifier and includes fitting the vectorizer or
building the token vocabulary. Inference time is wall-clock time over the
test split including feature extraction or encoding, taken as the fastest of
three passes to reduce scheduling jitter, and throughput is the test split
size divided by that time. Size is the serialized byte length of everything
required at inference, that is the vectorizer together with the classifier
for the logistic and MLP models, and the state dictionary together
with the token vocabulary for the convolutional and recurrent models.
Parameter counts are the fitted coefficients for the logistic models and the
total element count of all tensors for the others, so they are not comparable
between the two families.

Table~\ref{tab:cost} reports medians rather than means because training time
is right-skewed: a model that is slow on one category and seed drags its mean
without being representative. The first call into a deep-learning framework
in a process also pays one-off initialisation unrelated to the model. The
released script measures that directly before warming up: on the first
category and seed the convolutional model took 4.35 seconds cold against 2.64
seconds for the identical fit immediately afterwards, so roughly 1.7 seconds
of the first measurement is framework start-up. Both families are therefore
exercised once on a small throwaway problem before any measurement is taken,
and both the warm-up and the cold probe are part of the released script.

All ratios quoted in Section~\ref{sec:cost} are computed from the unrounded
medians in the released results, not from the rounded values printed in
Table~\ref{tab:cost}, so dividing two printed figures will not always
reproduce the last digit.

Earlier versions of this benchmark timed each fit with Python memory tracing
(\texttt{tracemalloc}) switched on. Tracing hooks every allocation and
inflated the logistic-model fit times in particular; this largely explains
why those versions reported the CNN and LSTM at 65 and 77 times the unigram model's
training time rather than the 212 and 258 times reported here. The timed fit
now runs with tracing off, and peak heap allocation is measured in a second,
identical fit that is not timed. It is in the released results but not in
Table~\ref{tab:cost}, because it undercounts tensor memory held outside the
Python allocator and so is not comparable across the two families.

All measurements were taken in a single session on one CPU thread of an Intel
Xeon at 2.80\,GHz, with the BLAS and OpenMP thread pools limited to one through
\pkg{threadpoolctl} and environment variables, and PyTorch set to one
intra-op and one inter-op thread; the thread-pool state during measurement is
recorded in \path{results/cost_benchmark_env.json}. Models are therefore
compared on arithmetic performed rather than on parallelism obtained. The test
split is 330 items, so the reported throughput includes per-call overhead that
a larger batch amortises, and amortises unevenly. Measured on the unigram and
character models over all six categories at 3{,}300, 33{,}000 and 330{,}000
items, the unigram model gains roughly half its throughput (a median factor
of 1.50 to 1.58) while the character model gains 4 to 9\%, so the median ratio
between them widens from 4.42 at 330 items to 6.10, 6.79 and 6.32 at the
three larger batches rather than staying put. Per-batch figures are in
\path{results/throughput_by_batch.csv} and its summary file. The host is
virtualised and shared, so a repeat run reproduces the ratios rather than the
absolute rates.

\section{Paired comparisons}
\label{app:pairs}

Table~\ref{tab:pairs} lists every paired comparison behind the bands of
Section~\ref{sec:protocol}: the 21 pairs of models trained from scratch and
the 24 distinct pairs involving a pretrained encoder
(\path{results/pairwise_bands.csv}, written by \path{src/pairwise_bands.py}
from \path{results/significance_pairwise.csv} and
\path{results/pretrained/pretrained_pairwise.csv}). Pretrained-encoder $F_1$
values are compared after rounding to four decimal places, as in
Table~\ref{tab:pretrained}. The Holm adjustment is applied separately within
each of the two families. The $p$-values are descriptive only: the 30 cells
are repeated splits of six corpora from one generator, not independent
replications, so they overstate the evidence, and no conclusion in the text
rests on them.

\begin{table}[p]
\caption{\label{tab:pairs}Every paired comparison over the 30
category--seed cells, oriented on the model with the higher mean $F_1$
(``Leader''). W--L--T counts the cells the leader wins, loses and ties;
``Share'' is the percentage of non-tied cells it wins; ``Diff.'' is the mean
difference in $F_1$ points; the band follows Section~\ref{sec:protocol}
(indist.\ = indistinguishable). Wilcoxon signed-rank $p$ is two-sided and
Holm-adjusted within each family. Descriptive only, because the cells are
not independent.}
\begin{center}\footnotesize
\setlength{\tabcolsep}{4pt}
\begin{tabular}{llcrrlll}
\toprule
Leader & Other & W--L--T & Share & Diff. & Band & Wilcoxon $p$ & Holm $p$ \\
\midrule
\multicolumn{8}{l}{\emph{Models trained from scratch}} \\
Char $n$-gram & LSTM & $30$--$0$--$0$ & $100$ & $3.98$ & reliable & $1.7\times10^{-6}$ & $3.3\times10^{-5}$ \\
Char $n$-gram & CNN & $30$--$0$--$0$ & $100$ & $3.49$ & reliable & $1.9\times10^{-9}$ & $3.9\times10^{-8}$ \\
Char $n$-gram & Word $1$--$3$-gram & $30$--$0$--$0$ & $100$ & $3.25$ & reliable & $1.7\times10^{-6}$ & $3.3\times10^{-5}$ \\
Char $n$-gram & Unigram & $30$--$0$--$0$ & $100$ & $3.11$ & reliable & $1.7\times10^{-6}$ & $3.3\times10^{-5}$ \\
Char $n$-gram & Word $1$--$2$-gram & $30$--$0$--$0$ & $100$ & $3.01$ & reliable & $1.7\times10^{-6}$ & $3.3\times10^{-5}$ \\
Char $n$-gram & MLP & $30$--$0$--$0$ & $100$ & $2.03$ & reliable & $1.7\times10^{-6}$ & $3.3\times10^{-5}$ \\
Word $1$--$2$-gram & Word $1$--$3$-gram & $17$--$0$--$13$ & $100$ & $0.24$ & reliable & $2.9\times10^{-4}$ & $3.2\times10^{-3}$ \\
MLP & LSTM & $28$--$2$--$0$ & $93$ & $1.95$ & reliable & $5.2\times10^{-6}$ & $7.3\times10^{-5}$ \\
MLP & CNN & $27$--$3$--$0$ & $90$ & $1.46$ & reliable & $3.9\times10^{-7}$ & $7.7\times10^{-6}$ \\
MLP & Word $1$--$3$-gram & $25$--$5$--$0$ & $83$ & $1.22$ & small, consistent & $1.8\times10^{-5}$ & $2.4\times10^{-4}$ \\
MLP & Unigram & $25$--$5$--$0$ & $83$ & $1.08$ & small, consistent & $4.1\times10^{-5}$ & $4.9\times10^{-4}$ \\
MLP & Word $1$--$2$-gram & $23$--$7$--$0$ & $77$ & $0.98$ & small, consistent & $4.5\times10^{-4}$ & $4.5\times10^{-3}$ \\
CNN & LSTM & $21$--$8$--$1$ & $72$ & $0.49$ & small, consistent & $0.02$ & $0.13$ \\
Word $1$--$2$-gram & LSTM & $20$--$9$--$1$ & $69$ & $0.97$ & indist. & $7.6\times10^{-3}$ & $0.07$ \\
Unigram & Word $1$--$3$-gram & $17$--$9$--$4$ & $65$ & $0.14$ & indist. & $0.16$ & $0.49$ \\
Unigram & CNN & $18$--$10$--$2$ & $64$ & $0.38$ & indist. & $0.12$ & $0.49$ \\
Unigram & LSTM & $19$--$11$--$0$ & $63$ & $0.87$ & indist. & $0.02$ & $0.13$ \\
Word $1$--$3$-gram & LSTM & $18$--$11$--$1$ & $62$ & $0.73$ & indist. & $0.03$ & $0.17$ \\
Word $1$--$2$-gram & CNN & $17$--$12$--$1$ & $59$ & $0.48$ & indist. & $0.06$ & $0.32$ \\
Word $1$--$3$-gram & CNN & $17$--$12$--$1$ & $59$ & $0.24$ & indist. & $0.38$ & $0.76$ \\
Word $1$--$2$-gram & Unigram & $12$--$10$--$8$ & $55$ & $0.10$ & indist. & $0.51$ & $0.76$ \\
\midrule
\multicolumn{8}{l}{\emph{Pretrained encoders, against the seven models and each other}} \\
DistilBERT, matched & LSTM & $30$--$0$--$0$ & $100$ & $3.59$ & reliable & $1.7\times10^{-6}$ & $4.0\times10^{-5}$ \\
DistilBERT, 3 ep. & LSTM & $30$--$0$--$0$ & $100$ & $3.11$ & reliable & $1.9\times10^{-9}$ & $4.5\times10^{-8}$ \\
DistilBERT, matched & CNN & $30$--$0$--$0$ & $100$ & $3.10$ & reliable & $1.7\times10^{-6}$ & $4.0\times10^{-5}$ \\
DistilBERT, matched & Word $1$--$3$-gram & $30$--$0$--$0$ & $100$ & $2.86$ & reliable & $1.7\times10^{-6}$ & $4.0\times10^{-5}$ \\
Char $n$-gram & MiniLM & $30$--$0$--$0$ & $100$ & $2.75$ & reliable & $1.7\times10^{-6}$ & $4.0\times10^{-5}$ \\
DistilBERT, matched & Unigram & $30$--$0$--$0$ & $100$ & $2.72$ & reliable & $1.7\times10^{-6}$ & $4.0\times10^{-5}$ \\
DistilBERT, matched & Word $1$--$2$-gram & $30$--$0$--$0$ & $100$ & $2.62$ & reliable & $1.7\times10^{-6}$ & $4.0\times10^{-5}$ \\
DistilBERT, 3 ep. & CNN & $29$--$0$--$1$ & $100$ & $2.62$ & reliable & $2.6\times10^{-6}$ & $4.4\times10^{-5}$ \\
DistilBERT, 3 ep. & Word $1$--$3$-gram & $29$--$0$--$1$ & $100$ & $2.38$ & reliable & $2.6\times10^{-6}$ & $4.4\times10^{-5}$ \\
DistilBERT, 3 ep. & Unigram & $29$--$0$--$1$ & $100$ & $2.25$ & reliable & $2.6\times10^{-6}$ & $4.4\times10^{-5}$ \\
DistilBERT, 3 ep. & Word $1$--$2$-gram & $29$--$0$--$1$ & $100$ & $2.14$ & reliable & $2.6\times10^{-6}$ & $4.4\times10^{-5}$ \\
DistilBERT, matched & MiniLM & $29$--$1$--$0$ & $97$ & $2.36$ & reliable & $2.9\times10^{-6}$ & $4.4\times10^{-5}$ \\
DistilBERT, matched & MLP & $28$--$1$--$1$ & $97$ & $1.64$ & reliable & $4.3\times10^{-6}$ & $5.2\times10^{-5}$ \\
DistilBERT, 3 ep. & MiniLM & $28$--$2$--$0$ & $93$ & $1.89$ & reliable & $2.4\times10^{-5}$ & $2.6\times10^{-4}$ \\
Char $n$-gram & DistilBERT, 3 ep. & $23$--$2$--$5$ & $92$ & $0.86$ & reliable & $2.5\times10^{-5}$ & $2.6\times10^{-4}$ \\
DistilBERT, matched & DistilBERT, 3 ep. & $21$--$2$--$7$ & $91$ & $0.48$ & reliable & $1.1\times10^{-4}$ & $8.9\times10^{-4}$ \\
DistilBERT, 3 ep. & MLP & $24$--$4$--$2$ & $86$ & $1.17$ & small, consistent & $9.8\times10^{-5}$ & $8.9\times10^{-4}$ \\
Char $n$-gram & DistilBERT, matched & $18$--$3$--$9$ & $86$ & $0.39$ & small, consistent & $1.1\times10^{-3}$ & $6.6\times10^{-3}$ \\
MiniLM & LSTM & $25$--$5$--$0$ & $83$ & $1.23$ & small, consistent & $2.9\times10^{-4}$ & $2.1\times10^{-3}$ \\
MLP & MiniLM & $22$--$8$--$0$ & $73$ & $0.72$ & small, consistent & $0.01$ & $0.06$ \\
MiniLM & CNN & $20$--$9$--$1$ & $69$ & $0.73$ & indist. & $0.03$ & $0.11$ \\
MiniLM & Word $1$--$3$-gram & $17$--$12$--$1$ & $59$ & $0.50$ & indist. & $0.14$ & $0.43$ \\
MiniLM & Unigram & $17$--$12$--$1$ & $59$ & $0.36$ & indist. & $0.29$ & $0.59$ \\
MiniLM & Word $1$--$2$-gram & $16$--$13$--$1$ & $55$ & $0.26$ & indist. & $0.42$ & $0.59$ \\
\bottomrule
\end{tabular}
\end{center}
\end{table}

\clearpage
\bibliography{references}

\begin{thebibliography}{51}
\providecommand{\natexlab}[1]{#1}
\providecommand{\url}[1]{\texttt{#1}}
\expandafter\ifx\csname urlstyle\endcsname\relax
  \providecommand{\doi}[1]{doi: #1}\else
  \providecommand{\doi}{doi: \begingroup \urlstyle{rm}\Url}\fi

\bibitem[Benavoli et~al.(2017)Benavoli, Corani, Dem{\v{s}}ar, and
  Zaffalon]{Benavoli2017}
Alessio Benavoli, Giorgio Corani, Janez Dem{\v{s}}ar, and Marco Zaffalon.
\newblock Time for a change: a tutorial for comparing multiple classifiers
  through {Bayesian} analysis.
\newblock \emph{Journal of Machine Learning Research}, 18\penalty0
  (77):\penalty0 1--36, 2017.
\newblock URL \url{https://jmlr.org/papers/v18/16-305.html}.

\bibitem[Ber{\k{e}}sewicz et~al.(2024)Ber{\k{e}}sewicz, Wydmuch, Cherniaiev,
  and Pater]{Beresewicz2024}
Maciej Ber{\k{e}}sewicz, Marek Wydmuch, Herman Cherniaiev, and Robert Pater.
\newblock Multilingual hierarchical classification of job advertisements for
  job vacancy statistics.
\newblock Technical Report arXiv:2411.03779, arXiv, 2024.

\bibitem[Berki et~al.(2025)Berki, Andicsov{\'a}, and Oravec]{Berki2025}
Martin Berki, Viera Andicsov{\'a}, and Milo{\v{s}} Oravec.
\newblock {NLP}-enhanced inflation measurement using {BERT} and web scraping.
\newblock \emph{Frontiers in Artificial Intelligence}, 8, 2025.
\newblock \doi{10.3389/frai.2025.1520659}.

\bibitem[Beskorovainyi(2026)]{Beskorovainyi2026arxiv}
Vladimir Beskorovainyi.
\newblock Machine learning for coding retail product names to consumer-price
  categories: A rule-plus-bag-of-words pipeline with reliability-weighted
  human-in-the-loop labeling, 2026.
\newblock URL \url{https://arxiv.org/abs/2606.02004}.

\bibitem[Bia{\l}ek and Ber{\k{e}}sewicz(2021)]{Bialek2021}
Jacek Bia{\l}ek and Maciej Ber{\k{e}}sewicz.
\newblock Scanner data in inflation measurement: From raw data to price
  indices.
\newblock \emph{Statistical Journal of the IAOS}, 37\penalty0 (4):\penalty0
  1315--1336, 2021.
\newblock \doi{10.3233/SJI-210816}.

\bibitem[Bojanowski et~al.(2017)Bojanowski, Grave, Joulin, and
  Mikolov]{Bojanowski2017}
Piotr Bojanowski, Edouard Grave, Armand Joulin, and Tomas Mikolov.
\newblock Enriching word vectors with subword information.
\newblock \emph{Transactions of the Association for Computational Linguistics},
  5:\penalty0 135--146, 2017.
\newblock \doi{10.1162/tacl_a_00051}.

\bibitem[Cavallo(2017)]{Cavallo2017}
Alberto Cavallo.
\newblock Are online and offline prices similar? {E}vidence from large
  multi-channel retailers.
\newblock \emph{American Economic Review}, 107\penalty0 (1):\penalty0 283--303,
  2017.
\newblock \doi{10.1257/aer.20160542}.

\bibitem[Cavallo and Rigobon(2016)]{CavalloRigobon2016}
Alberto Cavallo and Roberto Rigobon.
\newblock The {Billion Prices Project}: Using online prices for measurement and
  research.
\newblock \emph{Journal of Economic Perspectives}, 30\penalty0 (2):\penalty0
  151--178, 2016.
\newblock \doi{10.1257/jep.30.2.151}.

\bibitem[Chessa(2016)]{Chessa2016}
Antonio~G. Chessa.
\newblock A new methodology for processing scanner data in the {Dutch CPI}.
\newblock \emph{EURONA --- Eurostat Review on National Accounts and
  Macroeconomic Indicators}, 1/2016:\penalty0 49--69, 2016.
\newblock URL
  \url{https://ec.europa.eu/eurostat/web/products-statistical-books/-/ks-gp-16-001}.
\newblock Catalogue KS-GP-16-001-EN-N.

\bibitem[Dawid and Skene(1979)]{DawidSkene1979}
A.~P. Dawid and A.~M. Skene.
\newblock Maximum likelihood estimation of observer error-rates using the {EM}
  algorithm.
\newblock \emph{Journal of the Royal Statistical Society Series C: Applied
  Statistics}, 28\penalty0 (1):\penalty0 20--28, 1979.
\newblock \doi{10.2307/2346806}.

\bibitem[de~Haan and van~der Grient(2011)]{deHaanGrient2011}
Jan de~Haan and Heymerik~A. van~der Grient.
\newblock Eliminating chain drift in price indexes based on scanner data.
\newblock \emph{Journal of Econometrics}, 161\penalty0 (1):\penalty0 36--46,
  2011.
\newblock \doi{10.1016/j.jeconom.2010.09.004}.

\bibitem[Dem{\v{s}}ar(2006)]{Demsar2006}
Janez Dem{\v{s}}ar.
\newblock Statistical comparisons of classifiers over multiple data sets.
\newblock \emph{Journal of Machine Learning Research}, 7\penalty0 (1):\penalty0
  1--30, 2006.
\newblock URL \url{https://jmlr.org/papers/v7/demsar06a.html}.

\bibitem[Devlin et~al.(2019)Devlin, Chang, Lee, and Toutanova]{Devlin2019}
Jacob Devlin, Ming-Wei Chang, Kenton Lee, and Kristina Toutanova.
\newblock {BERT}: Pre-training of deep bidirectional transformers for language
  understanding.
\newblock In \emph{Proceedings of the 2019 Conference of the North American
  Chapter of the Association for Computational Linguistics: Human Language
  Technologies, Volume 1 (Long and Short Papers)}, pages 4171--4186, 2019.
\newblock \doi{10.18653/v1/N19-1423}.

\bibitem[Dietterich(1998)]{Dietterich1998}
Thomas~G. Dietterich.
\newblock Approximate statistical tests for comparing supervised classification
  learning algorithms.
\newblock \emph{Neural Computation}, 10\penalty0 (7):\penalty0 1895--1923,
  1998.
\newblock \doi{10.1162/089976698300017197}.

\bibitem[{Eurostat}(2017)]{EurostatScanner2017}
{Eurostat}.
\newblock Harmonised index of consumer prices: Practical guide for processing
  supermarket scanner data.
\newblock Technical report, European Commission, Eurostat, Unit C-4,
  Luxembourg, May 2017.
\newblock URL \url{https://stats.unece.org/ottawagroup/download/f477.pdf}.
\newblock Draft.

\bibitem[Harms and Spinder(2019)]{HarmsSpinder2019}
Alexander Harms and Siemen Spinder.
\newblock A comprehensive view of machine learning techniques for {CPI}
  production.
\newblock Discussion paper, Statistics Netherlands (CBS), The Hague, November
  2019.
\newblock URL
  \url{https://www.cbs.nl/-/media/_pdf/2019/47/machine-learning-for-cpi-production-2019.pdf}.

\bibitem[Harris et~al.(2020)Harris, Millman, van~der Walt, Gommers, Virtanen,
  Cournapeau, Wieser, Taylor, Berg, Smith, Kern, Picus, Hoyer, van Kerkwijk,
  Brett, Haldane, del R{\'i}o, Wiebe, Peterson, G{\'e}rard-Marchant, Sheppard,
  Reddy, Weckesser, Abbasi, Gohlke, and Oliphant]{numpy}
Charles~R. Harris, K.~Jarrod Millman, St{\'e}fan~J. van~der Walt, Ralf Gommers,
  Pauli Virtanen, David Cournapeau, Eric Wieser, Julian Taylor, Sebastian Berg,
  Nathaniel~J. Smith, Robert Kern, Matti Picus, Stephan Hoyer, Marten~H. van
  Kerkwijk, Matthew Brett, Allan Haldane, Jaime~Fern{\'a}ndez del R{\'i}o, Mark
  Wiebe, Pearu Peterson, Pierre G{\'e}rard-Marchant, Kevin Sheppard, Tyler
  Reddy, Warren Weckesser, Hameer Abbasi, Christoph Gohlke, and Travis~E.
  Oliphant.
\newblock Array programming with {NumPy}.
\newblock \emph{Nature}, 585:\penalty0 357--362, 2020.
\newblock \doi{10.1038/s41586-020-2649-2}.

\bibitem[Holm(1979)]{Holm1979}
Sture Holm.
\newblock A simple sequentially rejective multiple test procedure.
\newblock \emph{Scandinavian Journal of Statistics}, 6\penalty0 (2):\penalty0
  65--70, 1979.
\newblock URL \url{https://www.jstor.org/stable/4615733}.

\bibitem[{International Labour Office} et~al.(2020){International Labour
  Office}, {International Monetary Fund}, {Organisation for Economic
  Co-operation and Development}, {Eurostat}, {United Nations Economic
  Commission for Europe}, and {The World Bank}]{CPIManual}
{International Labour Office}, {International Monetary Fund}, {Organisation for
  Economic Co-operation and Development}, {Eurostat}, {United Nations Economic
  Commission for Europe}, and {The World Bank}.
\newblock \emph{Consumer Price Index Manual, 2020: Concepts and Methods}.
\newblock International Monetary Fund, Washington, DC, 2020.
\newblock \doi{10.5089/9781484354841.069}.

\bibitem[Jentoft et~al.(2022)Jentoft, Toth, and M{\"u}ller]{Jentoft2022}
Susie Jentoft, Bernhard Toth, and Diana M{\"u}ller.
\newblock From manual to machine: Challenges in machine learning for {COICOP}
  coding.
\newblock In \emph{Proceedings of the Nordic Statistical Meeting},
  Reykjav{\'i}k, Iceland, 2022.
\newblock URL
  \url{https://static1.squarespace.com/static/606f36b890215d7048ddaac0/t/62ed23ac0fefd90598690723/1659708333795/FROM+MANUAL+TO+MACHINE+-+CHALLENGES+IN+MACHINE+LEARNING+FOR+COICOP+CODING.pdf}.

\bibitem[Joulin et~al.(2017)Joulin, Grave, Bojanowski, and Mikolov]{Joulin2017}
Armand Joulin, Edouard Grave, Piotr Bojanowski, and Tomas Mikolov.
\newblock Bag of tricks for efficient text classification.
\newblock In \emph{Proceedings of the 15th Conference of the European Chapter
  of the Association for Computational Linguistics: Volume 2, Short Papers},
  pages 427--431, 2017.
\newblock \doi{10.18653/v1/E17-2068}.

\bibitem[Loshchilov and Hutter(2019)]{Loshchilov2019}
Ilya Loshchilov and Frank Hutter.
\newblock Decoupled weight decay regularization.
\newblock In \emph{International Conference on Learning Representations
  (ICLR)}, 2019.

\bibitem[Martindale et~al.(2020)Martindale, Rowland, Flower, and
  Clews]{Martindale2020}
Hazel Martindale, Edward Rowland, Tanya Flower, and Gareth Clews.
\newblock Semi-supervised machine learning with word embedding for
  classification in price statistics.
\newblock \emph{Data \& Policy}, 2:\penalty0 e12, 2020.
\newblock \doi{10.1017/dap.2020.13}.

\bibitem[McKinney(2010)]{pandas}
Wes McKinney.
\newblock Data structures for statistical computing in {Python}.
\newblock \emph{Proceedings of the 9th Python in Science Conference}, pages
  56--61, 2010.
\newblock \doi{10.25080/Majora-92bf1922-00a}.

\bibitem[Meertens et~al.(2022)Meertens, Diks, van~den Herik, and
  Takes]{Meertens2022}
Q.~A. Meertens, C.~G.~H. Diks, H.~J. van~den Herik, and F.~W. Takes.
\newblock Improving the output quality of official statistics based on machine
  learning algorithms.
\newblock \emph{Journal of Official Statistics}, 38\penalty0 (2):\penalty0
  485--508, 2022.
\newblock \doi{10.2478/jos-2022-0023}.

\bibitem[Mikolov et~al.(2013)Mikolov, Chen, Corrado, and Dean]{Mikolov2013}
Tomas Mikolov, Kai Chen, Greg Corrado, and Jeffrey Dean.
\newblock Efficient estimation of word representations in vector space.
\newblock Technical Report arXiv:1301.3781, arXiv, 2013.

\bibitem[Molladavoudi and Yung(2023)]{Molladavoudi2023}
Saeid Molladavoudi and Wesley Yung.
\newblock Exploring quality dimensions in trustworthy machine learning in the
  context of official statistics: Model explainability and uncertainty
  quantification.
\newblock \emph{AStA Wirtschafts- und Sozialstatistisches Archiv}, 17\penalty0
  (3):\penalty0 223--252, 2023.
\newblock \doi{10.1007/s11943-023-00331-z}.

\bibitem[Montbroussous and Monziols(2024)]{Montbroussous2024}
Adrien Montbroussous and Mathieu Monziols.
\newblock Classification of scanner data into {COICOP}: A machine learning
  approach.
\newblock In \emph{Proceedings of the 18th Meeting of the Ottawa Group on Price
  Indices}, Ottawa, Canada, 2024.
\newblock URL
  \url{https://stats.unece.org/ottawagroup/download/Papers-poster-Classification-of-scanner-data-into-COICOP.pdf}.

\bibitem[Nadeau and Bengio(2003)]{NadeauBengio2003}
Claude Nadeau and Yoshua Bengio.
\newblock Inference for the generalization error.
\newblock \emph{Machine Learning}, 52\penalty0 (3):\penalty0 239--281, 2003.
\newblock \doi{10.1023/A:1024068626366}.

\bibitem[Oancea(2023)]{Oancea2023}
Bogdan Oancea.
\newblock Automatic product classification using supervised machine learning
  algorithms in price statistics.
\newblock \emph{Mathematics}, 11\penalty0 (7):\penalty0 1588, 2023.
\newblock \doi{10.3390/math11071588}.

\bibitem[{Office for National Statistics}(2020)]{ONSscanner}
{Office for National Statistics}.
\newblock Automated classification of web-scraped clothing data in consumer
  price statistics.
\newblock Technical report, Office for National Statistics, Newport, UK,
  September 2020.
\newblock URL
  \url{https://www.ons.gov.uk/economy/inflationandpriceindices/articles/automatedclassificationofwebscrapedclothingdatainconsumerpricestatistics/2020-09-01}.

\bibitem[{Office for National Statistics}(2021)]{ONSnewdata2021}
{Office for National Statistics}.
\newblock Classification of new data in {UK} consumer price statistics.
\newblock Technical report, Office for National Statistics, Newport, UK, April
  2021.
\newblock URL
  \url{https://www.ons.gov.uk/economy/inflationandpriceindices/articles/classificationofnewdatainukconsumerpricestatistics/2021-04-06}.

\bibitem[Paszke et~al.(2019)Paszke, Gross, Massa, Lerer, Bradbury, Chanan,
  Killeen, Lin, Gimelshein, Antiga, Desmaison, K{\"o}pf, Yang, DeVito, Raison,
  Tejani, Chilamkurthy, Steiner, Fang, Bai, and Chintala]{pytorch}
Adam Paszke, Sam Gross, Francisco Massa, Adam Lerer, James Bradbury, Gregory
  Chanan, Trevor Killeen, Zeming Lin, Natalia Gimelshein, Luca Antiga, Alban
  Desmaison, Andreas K{\"o}pf, Edward Yang, Zachary DeVito, Martin Raison,
  Alykhan Tejani, Sasank Chilamkurthy, Benoit Steiner, Lu~Fang, Junjie Bai, and
  Soumith Chintala.
\newblock {PyTorch}: An imperative style, high-performance deep learning
  library.
\newblock In \emph{Advances in Neural Information Processing Systems 32}, pages
  8024--8035, 2019.
\newblock \doi{10.48550/arXiv.1912.01703}.

\bibitem[Pedregosa et~al.(2011)Pedregosa, Varoquaux, Gramfort, Michel, Thirion,
  Grisel, Blondel, Prettenhofer, Weiss, Dubourg, Vanderplas, Passos,
  Cournapeau, Brucher, Perrot, and Duchesnay]{sklearn}
Fabian Pedregosa, Ga{\"e}l Varoquaux, Alexandre Gramfort, Vincent Michel,
  Bertrand Thirion, Olivier Grisel, Mathieu Blondel, Peter Prettenhofer, Ron
  Weiss, Vincent Dubourg, Jake Vanderplas, Alexandre Passos, David Cournapeau,
  Matthieu Brucher, Matthieu Perrot, and {\'E}douard Duchesnay.
\newblock Scikit-learn: Machine learning in {Python}.
\newblock \emph{Journal of Machine Learning Research}, 12:\penalty0 2825--2830,
  2011.
\newblock URL \url{https://www.jmlr.org/papers/v12/pedregosa11a.html}.

\bibitem[{Python Software Foundation}(2025)]{Python}
{Python Software Foundation}.
\newblock \emph{Python Language Reference, Version 3.13}.
\newblock Python Software Foundation, 2025.
\newblock URL \url{https://www.python.org/}.

\bibitem[Raykar et~al.(2010)Raykar, Yu, Zhao, Valadez, Florin, Bogoni, and
  Moy]{Raykar2010}
Vikas~C. Raykar, Shipeng Yu, Linda~H. Zhao, Gerardo~Hermosillo Valadez, Charles
  Florin, Luca Bogoni, and Linda Moy.
\newblock Learning from crowds.
\newblock \emph{Journal of Machine Learning Research}, 11\penalty0
  (43):\penalty0 1297--1322, 2010.
\newblock URL \url{https://jmlr.org/papers/v11/raykar10a.html}.

\bibitem[Reimers and Gurevych(2019)]{Reimers2019}
Nils Reimers and Iryna Gurevych.
\newblock Sentence-{BERT}: Sentence embeddings using {Siamese} {BERT}-networks.
\newblock In \emph{Proceedings of the 2019 Conference on Empirical Methods in
  Natural Language Processing and the 9th International Joint Conference on
  Natural Language Processing (EMNLP-IJCNLP)}, pages 3982--3992, 2019.

\bibitem[Ruocco(2024)]{AIML4OS2024}
Giulio Ruocco.
\newblock From text to code: Experiences and potential of the use of {AI/ML}
  for classifying and coding.
\newblock Technical report, ESSnet AI and Machine Learning for Official
  Statistics (AIML4OS), Work Package 10, 2024.
\newblock URL
  \url{https://cros.ec.europa.eu/book-page/aiml4os-wp10-text-code-experiences-and-potential-use-aiml-classifying-and-coding}.

\bibitem[Sanh et~al.(2019)Sanh, Debut, Chaumond, and Wolf]{Sanh2019}
Victor Sanh, Lysandre Debut, Julien Chaumond, and Thomas Wolf.
\newblock {DistilBERT}, a distilled version of {BERT}: smaller, faster, cheaper
  and lighter.
\newblock In \emph{5th Workshop on Energy Efficient Machine Learning and
  Cognitive Computing (EMC$^2$) at NeurIPS 2019}, 2019.
\newblock arXiv:1910.01108.

\bibitem[Schierholz and Schonlau(2021)]{Schierholz2021}
Malte Schierholz and Matthias Schonlau.
\newblock Machine learning for occupation coding---a comparison study.
\newblock \emph{Journal of Survey Statistics and Methodology}, 9\penalty0
  (5):\penalty0 1013--1034, 2021.
\newblock \doi{10.1093/jssam/smaa023}.

\bibitem[Sheng et~al.(2008)Sheng, Provost, and Ipeirotis]{Sheng2008}
Victor~S. Sheng, Foster Provost, and Panagiotis~G. Ipeirotis.
\newblock Get another label? {I}mproving data quality and data mining using
  multiple, noisy labelers.
\newblock In \emph{Proceedings of the 14th ACM SIGKDD International Conference
  on Knowledge Discovery and Data Mining}, pages 614--622, 2008.
\newblock \doi{10.1145/1401890.1401965}.

\bibitem[Snow et~al.(2008)Snow, O'Connor, Jurafsky, and Ng]{Snow2008}
Rion Snow, Brendan O'Connor, Daniel Jurafsky, and Andrew~Y. Ng.
\newblock Cheap and fast---but is it good? {E}valuating non-expert annotations
  for natural language tasks.
\newblock In \emph{Proceedings of the 2008 Conference on Empirical Methods in
  Natural Language Processing}, pages 254--263, 2008.
\newblock \doi{10.3115/1613715.1613751}.

\bibitem[Toth(2025)]{Toth2025}
Bernhard Toth.
\newblock Exploring the use of large language models for {COICOP}
  classification.
\newblock \emph{Statistical Journal of the IAOS}, 41\penalty0 (4), 2025.
\newblock \doi{10.1177/18747655251393372}.

\bibitem[{United Nations Statistics Division}(2023)]{COICOP}
{United Nations Statistics Division}.
\newblock Classification of individual consumption according to purpose
  ({COICOP}) 2018.
\newblock Statistical Papers Series M~99, United Nations, New York, 2023.
\newblock URL
  \url{https://unstats.un.org/unsd/classifications/Family/Detail/2094}.
\newblock Endorsed by the UN Statistical Commission at its 49th session, March
  2018.

\bibitem[Virtanen et~al.(2020)Virtanen, Gommers, Oliphant, Haberland, Reddy,
  Cournapeau, Burovski, Peterson, Weckesser, Bright, van~der Walt, Brett,
  Wilson, Millman, Mayorov, Nelson, Jones, Kern, Larson, Carey, Polat, Feng,
  Moore, VanderPlas, Laxalde, Perktold, Cimrman, Henriksen, Quintero, Harris,
  Archibald, Ribeiro, Pedregosa, and van Mulbregt]{scipy}
Pauli Virtanen, Ralf Gommers, Travis~E. Oliphant, Matt Haberland, Tyler Reddy,
  David Cournapeau, Evgeni Burovski, Pearu Peterson, Warren Weckesser, Jonathan
  Bright, St{\'e}fan~J. van~der Walt, Matthew Brett, Joshua Wilson, K.~Jarrod
  Millman, Nikolay Mayorov, Andrew R.~J. Nelson, Eric Jones, Robert Kern, Eric
  Larson, C.~J. Carey, Ilhan Polat, Yu~Feng, Eric~W. Moore, Jake VanderPlas,
  Denis Laxalde, Josef Perktold, Robert Cimrman, Ian Henriksen, E.~A. Quintero,
  Charles~R. Harris, Anne~M. Archibald, Ant{\^o}nio~H. Ribeiro, Fabian
  Pedregosa, and Paul van Mulbregt.
\newblock {SciPy} 1.0: Fundamental algorithms for scientific computing in
  {Python}.
\newblock \emph{Nature Methods}, 17:\penalty0 261--272, 2020.
\newblock \doi{10.1038/s41592-019-0686-2}.

\bibitem[Wang et~al.(2020)Wang, Wei, Dong, Bao, Yang, and Zhou]{Wang2020MiniLM}
Wenhui Wang, Furu Wei, Li~Dong, Hangbo Bao, Nan Yang, and Ming Zhou.
\newblock {MiniLM}: Deep self-attention distillation for task-agnostic
  compression of pre-trained transformers.
\newblock In \emph{Advances in Neural Information Processing Systems},
  volume~33, pages 5776--5788, 2020.

\bibitem[Whitehill et~al.(2009)Whitehill, Wu, Bergsma, Movellan, and
  Ruvolo]{Whitehill2009}
Jacob Whitehill, Ting-fan Wu, Jacob Bergsma, Javier~R. Movellan, and Paul~L.
  Ruvolo.
\newblock Whose vote should count more: Optimal integration of labels from
  labelers of unknown expertise.
\newblock In \emph{Advances in Neural Information Processing Systems 22 (NIPS
  2009)}, pages 2035--2043, 2009.
\newblock URL
  \url{https://papers.nips.cc/paper_files/paper/2009/hash/f899139df5e1059396431415e770c6dd-Abstract.html}.

\bibitem[Wilcoxon(1945)]{Wilcoxon1945}
Frank Wilcoxon.
\newblock Individual comparisons by ranking methods.
\newblock \emph{Biometrics Bulletin}, 1\penalty0 (6):\penalty0 80--83, 1945.
\newblock \doi{10.2307/3001968}.

\bibitem[Wolf et~al.(2020)Wolf, Debut, Sanh, Chaumond, Delangue, Moi, Cistac,
  Rault, Louf, Funtowicz, et~al.]{Wolf2020}
Thomas Wolf, Lysandre Debut, Victor Sanh, Julien Chaumond, Clement Delangue,
  Anthony Moi, Pierric Cistac, Tim Rault, R{\'e}mi Louf, Morgan Funtowicz,
  et~al.
\newblock Transformers: State-of-the-art natural language processing.
\newblock In \emph{Proceedings of the 2020 Conference on Empirical Methods in
  Natural Language Processing: System Demonstrations}, pages 38--45, 2020.

\bibitem[Yung et~al.(2022)Yung, Tam, Buelens, Chipman, Dumpert, Ascari, Rocci,
  Burger, and Choi]{Yung2022}
Wesley Yung, Siu-Ming Tam, Bart Buelens, Hugh Chipman, Florian Dumpert,
  Gabriele Ascari, Fabiana Rocci, Joep Burger, and Ia~Choi.
\newblock A quality framework for statistical algorithms.
\newblock \emph{Statistical Journal of the IAOS}, 38\penalty0 (1):\penalty0
  291--308, 2022.
\newblock \doi{10.3233/SJI-210875}.

\bibitem[Zhang et~al.(2015)Zhang, Zhao, and LeCun]{Zhang2015charcnn}
Xiang Zhang, Junbo Zhao, and Yann LeCun.
\newblock Character-level convolutional networks for text classification.
\newblock In \emph{Advances in Neural Information Processing Systems 28 (NIPS
  2015)}, pages 649--657, 2015.
\newblock URL
  \url{https://papers.nips.cc/paper/2015/hash/250cf8b51c773f3f8dc8b4be867a9a02-Abstract.html}.

\end{thebibliography}

\end{document}